\documentclass{sig-alternate}
\usepackage[utf8]{inputenc}
\usepackage{amssymb,amsmath,mathrsfs}
\usepackage{graphicx}
\usepackage[linesnumbered,ruled,lined]{algorithm2e}
\usepackage[font=small,labelfont=bf]{caption} 
\usepackage[small]{subfigure}
\usepackage{color}
\usepackage{url}
\usepackage{float}
\usepackage{multirow}
\usepackage{tabularx}
\usepackage{bbm}
\usepackage{enumitem}

\usepackage[table]{xcolor}

\newtheorem{definition}{Definition}

\newtheorem{hypothesis}{Hypothesis}

\newcommand{\nop}[1]{}
\def\ssp{\hspace*{0.4ex}}

\DeclareMathOperator{\R}{\mathbb{R}}
\DeclareMathOperator{\EE}{\mathbb{E}}

\DeclareMathOperator*{\argmax}{argmax}

\def\one{\mathbbm{1}}
\def\e{\mathrm{exp}}
\def\log{\mathrm{log}}
\DeclareMathOperator{\s}{\mathrm{s}}

\def\etal{{\em et al.\/}\,}

\def\ie{{\sl i.e.}}
\def\eg{{\sl e.g.}}

\def\O{\mathcal{O}}
\def\N{\mathcal{N}}
\def\F{\mathcal{F}}

\def\M{\mathcal{M}}
\def\Y{\mathcal{Y}}
\def\Yn{\overline{\mathcal{Y}}}
\def\D{\mathcal{D}}
\def\P{\mathcal{P}}
\def\C{\mathcal{C}}

\def\E{\mathcal{E}}
\def\T{\mathcal{T}}

\def\bu{\mathbf{u}}
\def\bc{\mathbf{c}}
\def\bv{\mathbf{v}}

\newcolumntype{x}{>{\hsize=.8\hsize}X}
\newcolumntype{a}{>{\hsize=.4\hsize}X}

\newfont{\mycrnotice}{ptmr8t at 7pt}
\newfont{\myconfname}{ptmri8t at 7pt}
\let\crnotice\mycrnotice
\let\confname\myconfname
\begin{document}

\title{Label Noise Reduction in Entity Typing by Heterogeneous Partial-Label Embedding}

\author{
\alignauthor
Xiang Ren$^{\dag *}$ $\quad$\ Wenqi He$^{\dag}\thanks{Equal contribution.}\quad$  Meng Qu$^{\dag}$ $\quad$Clare R. Voss$^{\ddagger}\quad$ Heng Ji$^{\sharp}\quad$ Jiawei Han$^{\dag}$\\[0.5ex]
\affaddr{$^{\dag}$ University of Illinois at Urbana-Champaign, Urbana, IL, USA}\\
\affaddr{$^{\ddagger}$ Computational \&  Information Sciences Directorate, Army Research Laboratory, Adelphi, MD, USA}\\
\affaddr{$^{\sharp}$ Computer Science Department, Rensselaer Polytechnic Institute, USA}\\
\email{\begin{footnotesize}$^{\dag}$\{xren7, wenqihe3, mengqu2, hanj\}@illinois.edu$\ssp$ $^{\ddagger}$clare.r.voss.civ@mail.mil$\ssp$ $^{\sharp}$jih@rpi.edu\end{footnotesize}}
}

\maketitle

\begin{abstract}
Current systems of fine-grained entity typing use distant supervision in conjunction with existing knowledge bases to assign categories (type labels) to entity mentions. However, the type labels so obtained from knowledge bases are often noisy (\ie, incorrect for the entity mention's local context). We define a \textit{new} task, \textit{Label Noise Reduction in Entity Typing} (LNR), to be the automatic identification of correct type labels (type-paths) for \textit{training examples}, given the set of candidate type labels obtained by distant supervision with a given type hierarchy. The unknown type labels for individual entity mentions and the semantic similarity between entity types pose unique challenges for solving the LNR task. We propose a general framework, called \textbf{\textsf{PLE}}, to \textit{jointly} embed entity mentions, text features and entity types into the same low-dimensional space where, in that space, \textit{objects whose types are semantically close  have similar representations}. Then we estimate the type-path for \textit{each} training example in a top-down manner using the learned embeddings. We formulate a global objective for learning the embeddings from text corpora and knowledge bases, which adopts a novel margin-based loss that is robust to \textit{noisy labels} and faithfully models \textit{type correlation} derived from knowledge bases. Our experiments on three public typing datasets demonstrate the effectiveness and robustness of PLE, with an average of 25\% improvement in accuracy compared to next best method.
\end{abstract}
\section{Introduction}
\label{sec:intro}
\begin{figure}
\centering
\begin{small}
\vspace{-0.1cm}
\includegraphics[width = 88 mm]{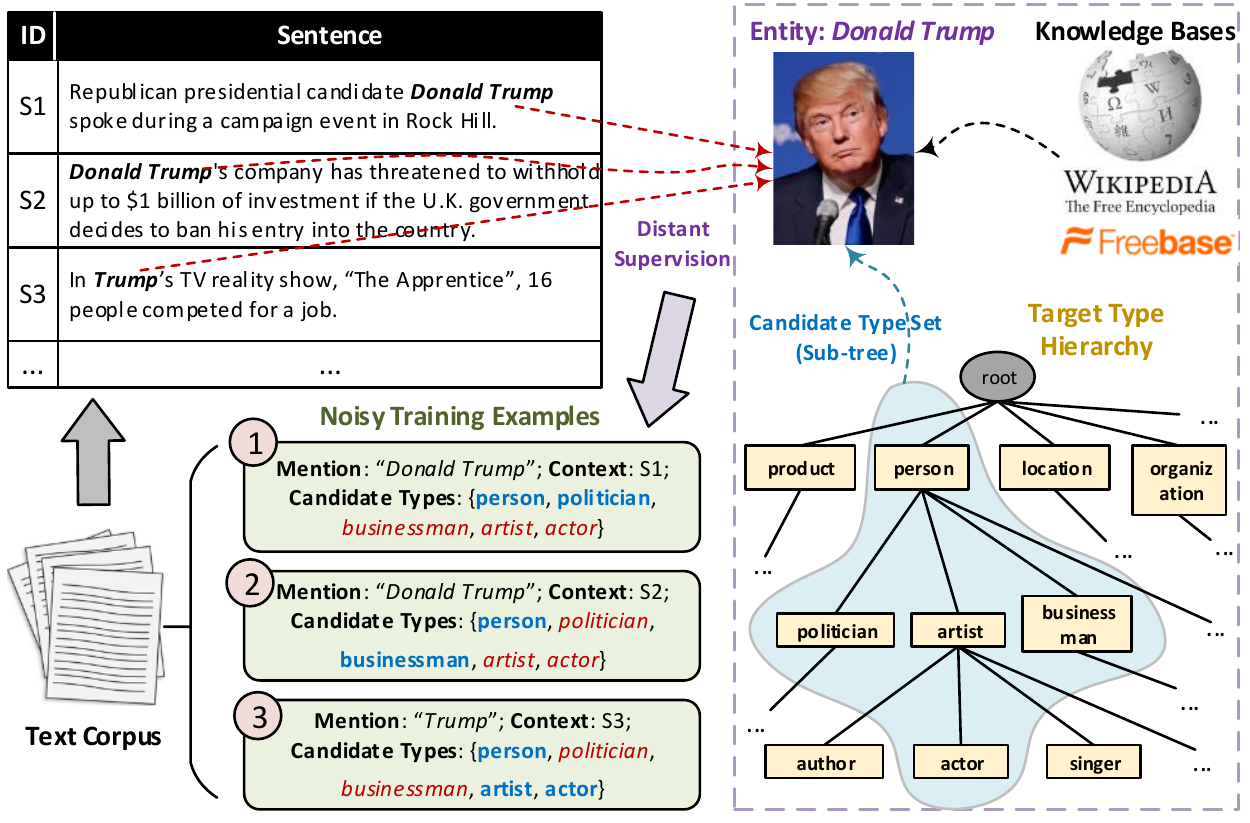}
\vspace{-0.5cm}
\caption{Current systems may find \textit{Donald Trump} mentioned in sentences S1-S3 and assign the same types to all (listed within braces), when only some types are correct for context (blue).}
\label{figure:motivated_example}
\end{small}
\vspace{-0.2cm}
\end{figure}

Entity typing is an important task in text analysis.
Assigning types (\eg, \begin{small}\texttt{person}, \texttt{location}, \texttt{organization}\end{small}) to mentions of entities in documents enables effective structured analysis of unstructured text corpora.
The extracted type information can be used in a wide range of ways (\eg, serving as primitives for information extraction \cite{schmitz2012open} and knowledge base (KB) completion~\cite{dong2014knowledge}, and assisting question answering~\cite{faderopenQA14}).
Traditional entity typing systems~\cite{ren2015clustype,nadeau2007survey} focus on a small set of coarse types (typically fewer than 10).
Recent studies~\cite{Yogatama2015embedding,ling2012fine,yosef2012hyena} work on a much larger set of fine-grained types which form a tree-structured hierarchy 
(\eg, \begin{small}\texttt{actor}\end{small} as a subtype of \begin{small}\texttt{artist}\end{small}, and \begin{small}\texttt{artist}\end{small} is a subtype of \begin{small}\texttt{person}\end{small}, as in blue region of Fig.~\ref{figure:motivated_example}).
While types are usually defined to be mutually exclusive within a coarse type set (\eg, by assuming a mention cannot be both \begin{small}\texttt{person}\end{small} and \begin{small}\texttt{location})\end{small},
fine-grained typing allows one mention to have multiple types, which together constitute one \textit{type-path} (not necessarily ending in a leaf node) in the given type hierarchy, \textit{depending on the local context} (\eg, sentence).
Consider the example in Fig.~\ref{figure:motivated_example}, ``\textit{Trump}" could be labeled as \begin{small}\{\texttt{person}, \texttt{artist}, \texttt{actor}\}\end{small} in S3 (TV show).
But he could also be labeled as \begin{small}\{\texttt{person}, \texttt{politician}\}\end{small} in S1 or \begin{small}\{\texttt{person}, \texttt{businessman}\}\end{small} in S2.

A major challenge in fine-grained typing is the absence of human-annotated data. The process of manually labeling a training set with large numbers of fine-grained types (usually over 100) is too expensive and error-prone (hard for annotators to distinguish over 100 types consistently).
Current systems annotate training corpora automatically using knowledge bases (\ie, \textit{distant supervision})~\cite{ren2015clustype,Yogatama2015embedding,ling2012fine,yosef2012hyena}.
A typical workflow of distant supervision is as follows (see Fig.~\ref{figure:motivated_example}):
(1) identify entity mentions in the documents;
(2) link mentions to entities in KB;
and (3) assign, to the candidate type set of each mention, all KB types of its KB-linked entity.
However, this approach introduces \textit{label noise} to the mentions since it fails to take the semantics of the mentions' local contexts into account when assigning type labels.
For example, in Fig.~\ref{figure:motivated_example}, the types assigned to entity \textit{Trump} include \begin{small}\texttt{person}, \texttt{artist}, \texttt{actor}, \texttt{politician}, \texttt{businessman}\end{small}, while only \begin{small}\{\texttt{person}, \texttt{politician}\}\end{small} are correct types for \textit{Trump} in S1.

Many previous studies ignore the label noise in automatically labeled training corpora---\textit{all} candidate types obtained by distant supervision are treated as ``true" types in training multi-label (hierarchical) classifiers~\cite{Yogatama2015embedding,ling2012fine,yosef2012hyena}.
This has become an impediment to improving the performance of current fine-grained typing systems as a majority of mentions in training sets have noisy types (see Table.~\ref{table:label_noise_stats}, row~(1)). A few systems try to denoise automatically labeled training corpora by simple pruning heuristics such as deleting mentions with conflicting types~\cite{gillick2014context}. However, such strategies significantly reduce the size of training set (Table~\ref{table:label_noise_stats}, rows (2a-c)) and lead to performance degradation (later shown in our experiments). The larger the target type set, the more severe the loss. So far there is no effective way to automatically create high-quality training data for fine-grained typing.

This motivated us to define a \textit{new} task: \emph{Label Noise Reduction in Entity Typing} (LNR), that is, identifying the correct type labels for \textit{each training example} from its noisy candidate type set (generated by distant supervision with a given type hierarchy).
While the typical entity typing systems assume that type labels in training data are all valid and focus on designing models to predict types for \textit{unlabeled mentions}, LNR focuses on identifying the correct types for \textit{automatically labeled mentions}, which is related to partial label learning~\cite{nguyen2008classification,cour2011learning}.
LNR is a fundamental task in building entity typing systems with distant supervision because it reduces the level of type label noise in the training data that, in turn, yields a better entity type classifier.

\begin{table}[t]
\vspace{-0.0cm}
\begin{scriptsize}
\begin{center}
\begin{tabularx}{\linewidth}{l|cccc}
\hline
\begin{scriptsize}
\textbf{Dataset}\end{scriptsize} & \begin{scriptsize}
\textbf{Wiki}\end{scriptsize}  & \begin{scriptsize}\textbf{OntoNotes}\end{scriptsize} & \begin{scriptsize}\textbf{BBN}\end{scriptsize} & \begin{scriptsize}\textbf{NYT}\end{scriptsize}\\ \hline
\# of target types & 113 & 89 & 47 & 446 \\
(1) noisy mentions (\%) & 27.99 & 25.94 & 22.32 & 51.81 \\
\hline
(2a) sibling pruning (\%)& 23.92 & 16.09 & 22.32 & 39.26 \\
(2b) min. pruning (\%)& 28.22 & 8.09 & 3.27 & 32.75 \\
(2c) all pruning (\%)& 45.99 & 23.45 & 25.33 & 61.12 \\
\hline
\end{tabularx}
\vspace{-0.2cm}
\caption{\textbf{A study of type label noise}. (1): \%mentions with multiple \textit{sibling types} (\eg, \texttt{actor}, \texttt{singer}); (2a)-(2c): \%mentions deleted by the three pruning heuristics~\cite{gillick2014context} (see Sec.~\ref{subsec:experiment_setting}), for three experiment datasets and New York Times annotation corpus~\cite{dunietz2014new}.}
\label{table:label_noise_stats}
\vspace{-0.3cm}
\end{center}
\end{scriptsize}
\end{table}

The presence of incorrect type labels in a mention's candidate type set poses a unique challenge to estimating the relatedness between entity mentions and types using fully/semi-supervised learning methods~\cite{Yogatama2015embedding,Dong2015HybridNeural,weston2011wsabie}---co-occurrence patterns alone between mentions and their candidate types in the corpus may be \textit{unreliable}, as shown in our example above.

We approach the LNR task as follows: (1) Model the true type labels in a candidate type set as latent variables and require only the ``\textit{best}" type (measured under the proposed metric) to be relevant to the mention---this requirement is less limiting compared with other multi-label learning methods that assume \textit{every} candidate type is relevant to the mention.
(2) Extract a variety of text features from entity mentions and their local contexts, and leverage corpus-level co-occurrences between mentions and features to model mentions' types.
(3) Model type correlation (semantic similarity) jointly with mention-candidate type associations and mention-feature co-occurrences, to assist type-path inference, by exploiting two signals: (i) the given type hierarchy, and (ii) the shared entities between two types in KB.

\begin{figure*}
\centering
\vspace{-0.6cm}
\includegraphics[width = 176 mm]{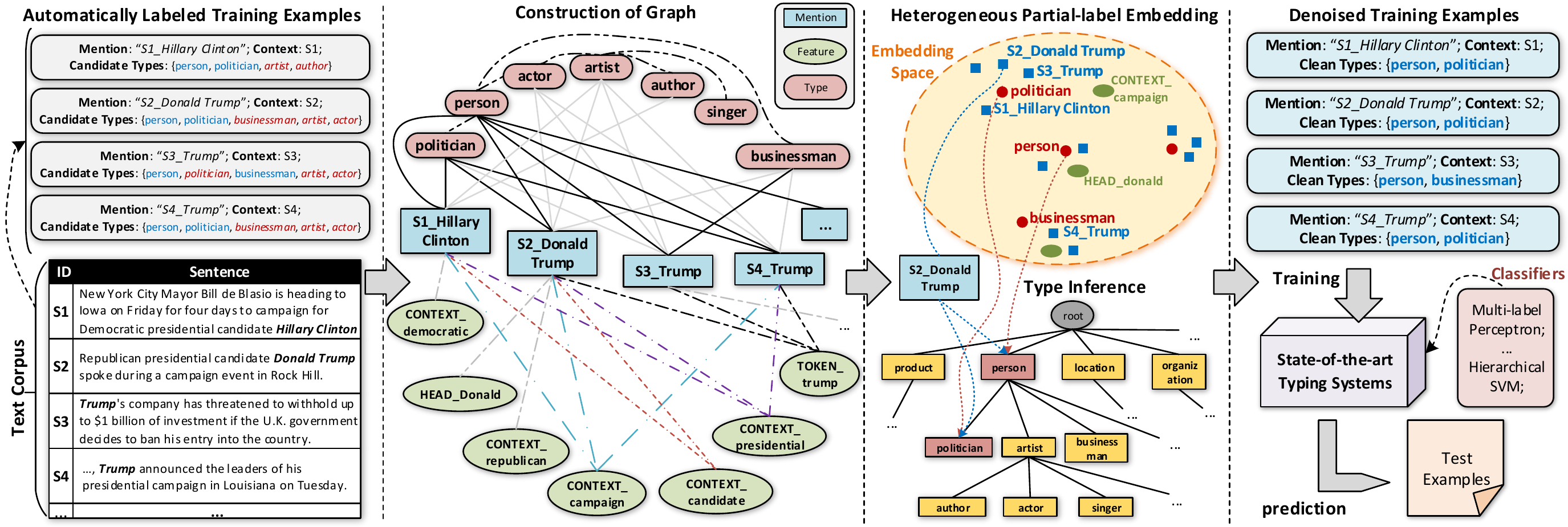}
\vspace{-0.3cm}
\caption{Framework Overview and Examples of Graph Construction.}
\label{figure:framework_overview}
\vspace{-0.4cm}
\end{figure*}

To integrate these elements of our approach, a principled framework, \textsf{Heterogeneous Partial-Label Embedding} (\textsf{PLE}), is proposed. First, PLE constructs a heterogeneous graph to represent three kinds of objects: entity mentions, text features and entity types, and their relationships in a unified form (see Fig.~\ref{figure:framework_overview}). Associations between mentions and their \textit{true} types are kept as latent structures in the graph to be estimated (Sec.~\ref{subsec:graph}). 
We formulate a global objective to jointly embed the graph into a low-dimensional space where, in that space, objects whose types are semantically close also have similar representations (see Sec.~\ref{subsec:model}).
Specifically, we design a novel margin-based rank loss to model \textit{mention-type associations}, which enforces only the \textit{best} candidate type to be embedded close to the mention (thus is robust to the \textit{false} candidate types). We further integrate the margin-based rank loss with the skip-gram model~\cite{mikolov2013distributed} to jointly capture the \textit{corpus-level mention-feature co-occurrences} and the \textit{KB-based type correlation} in the embedding process.
With the learned embeddings, we can efficiently estimate the correct type-path for each entity mention in the training set in a top-down manner. An efficient alternative minimization algorithm is developed to solve the optimization problem based on block-wise coordinate descent~\cite{tseng2001convergence} (see Sec.~\ref{subsec:algorithm}).
The major contributions of this paper are as follows:
\vspace{-0.1cm}
\begin{enumerate}[leftmargin=12pt]\itemsep+0.0cm
\item This is the first systematic study of noisy type labels in distant supervision. It defines a new task, \emph{Label Noise Reduction in Entity Typing}, to identify the correct type-path for each mention from its noisy candidate type set.
\item An embedding-based framework, PLE, is proposed.  It models and measures semantic similarity between entity mentions and type labels, and is robust to label noise.
\item A joint optimization problem is formulated that integrates mention-type association, corpus-level mention-fea-ture co-occurrence, and KB-based type correlation.
\item Experiments with three public fine-grained typing datasets demonstrate that PLE reduces their label noise substantially and, when PLE-denoised corpora are used as training sets, they also improve the performance of state-of-the-art fine-grained typing systems significantly.
\end{enumerate}
\vspace{-0.2cm}

\section{Background and Problem}
\label{subsec:problem}
\vspace{-0.0cm}
The input to LNR is a knowledge base \begin{small}$\Psi$\end{small} with type schema \begin{small}$\Y_{\Psi}$\end{small}, a target type hierarchy \begin{small}$\Y$\end{small} which covers a subset of types in \begin{small}$\Psi$\end{small}, \ie, \begin{small}$\Y\subseteq\Y_{\Psi}$\end{small}, and an automatically labeled training corpus \begin{small}$\D$\end{small} (obtained by distant supervision with \begin{small}$\Y$\end{small}).

\medskip
\noindent \textsf{\textbf{Knowledge Base and Target Type Hierarchy.}}
A KB with a set of entities \begin{small}$\E_{\Psi}$\end{small} contains human-curated facts on both entity-entity facts of various relationship types and entity-type facts. We denote \textit{entity-type facts} in a KB \begin{small}$\Psi$\end{small} (with type schema \begin{small}$\Y_{\Psi}$\end{small}) as \begin{small}$\T_\Psi = \big\{(e, y)\big\}\subset\E_{\Psi}\times \Y_{\Psi}$\end{small}.
A \textit{target type hierarchy} is a tree where nodes represent types of interests from \begin{small}$\Y_{\Psi}$\end{small} (or types which can be uniquely mapped to those in \begin{small}$\Y_{\Psi}$\end{small}). In existing entity typing studies, several fine-grained type hierarchies are manually/semi-automatically constructed using WordNet~\cite{yosef2012hyena} or Freebase~\cite{gillick2014context,ling2012fine}.

\medskip
\noindent \textsf{\textbf{Automatically Labeled Training Corpora.}}
Formally, a labeled corpus for entity typing consists of a set of extracted \textit{entity mentions} \begin{small}$\M=\{m_i\}_{i=1}^N$\end{small} (\ie, token spans representing entities in text), the \textit{context} (\eg, sentence, paragraph) of each mention \begin{small}$\{c_i\}_{i=1}^N$\end{small}, and the \textit{candidate type sets} \begin{small}$\{\Y_i\}_{i=1}^N$\end{small} automatically generated for each mention. We represent the training corpus using a set of mention-based triples \begin{small}$\D = \big\{(m_i, c_i, \Y_i)\big\}_{i=1}^N$\end{small}.
There exist publicly available automatically-labeled corpora such as Wikilinks~\cite{singh2012wikilinks} dataset where entity mentions have already been extracted and mapped to KB entities using anchor links in the corpus. In specific domains (\eg, customer reviews, tweets) where such public datasets are unavailable, one can utilize distant supervision~\cite{ren2015clustype,Dong2015HybridNeural,ling2012fine} to automatically label the corpus, where an entity linking system~\cite{shen2014entity} will detect mentions \begin{small}$m_i$\end{small} (in set \begin{small}$\M$\end{small}) and map them to one or more entity \begin{small}$e_i$\end{small} in \begin{small}$\E_\Psi$\end{small}.
Types of \begin{small}$e_i$\end{small} in KB \begin{small}$\Psi$\end{small} are then associated with \begin{small}$m_i$\end{small} to form its candidate type set \begin{small}$\Y_i$\end{small}, \ie, \begin{small}$\Y_i = \big\{y~|~(e_i, y)\in\T_\Psi,~y\in\Y\big\}$\end{small}.

\medskip
\noindent \textsf{\textbf{Problem Description.}}
Since \begin{small}$\Y_i$\end{small} is annotated for entity \begin{small}$e_i$\end{small}, it includes all possible types of \begin{small}$e_i$\end{small} and thus may contain types that are \textit{irrelevant} to \begin{small}$m_i$\end{small}'s specific context \begin{small}$c_i$\end{small}.
Ideally, the type labels for \begin{small}$m_i\in\M$\end{small} should form a \textit{type-path} (not required to end at a leaf) in \begin{small}$\Y_i$\end{small}~\cite{Yogatama2015embedding,gillick2014context,yosef2012hyena}, which serves as a \textit{context-dependent} type annotation for \begin{small}$m_i$\end{small}.
However, as discussed in~\cite{gillick2014context} and shown in Fig.~\ref{figure:motivated_example}, 
\begin{small}$\Y_i$\end{small} may contain type-paths that are irrelevant to \begin{small}$m_i$\end{small} in \begin{small}$c_i$\end{small}.
Even though in some cases \begin{small}$\Y_i$\end{small} is already a type-path, it may be overly specific for \begin{small}$c_i$\end{small} and so insufficient to infer the whole type-path using \begin{small}$c_i$\end{small}.
We denote the true type-path for mention \begin{small}$m_i$\end{small} as \begin{small}$\Y^*_i$\end{small}. This work focuses on estimating \begin{small}$\Y^*_i$\end{small} from \begin{small}$\Y_i$\end{small} based on mention \begin{small}$m_i$\end{small} as well as its context \begin{small}$c_i$\end{small}, where the candidate type set \begin{small}$\Y_i$\end{small} may contain (1) types that are irrelevant to \begin{small}$c_i$\end{small}, and (2) types that are overly specific to \begin{small}$c_i$\end{small}.
Formally, we define the LNR task as follows.
\begin{definition}[Problem Definition]
Given a KB \begin{small}$\Psi$\end{small} with type schema \begin{small}$\Y_\Psi$\end{small} and entity-type facts \begin{small}$\T_\Psi = \big\{(e, y)\big\}$\end{small}, a target type hierarchy \begin{small}$\Y\subseteq\Y_\Psi$\end{small}, and an automatically labeled training corpus \begin{small}$\D = \big\{(m_i, c_i, \Y_i)\big\}_{i=1}^N$\end{small}, the LNR task aims to estimate a single type-path \begin{small}$\Y^*_i\subseteq\Y_i$\end{small} for each entity mention \begin{small}$m_i\in\M$\end{small}, based on \begin{small}$m_i$\end{small} itself as well as its context \begin{small}$c_i$\end{small}.
\end{definition}
\medskip
\noindent \textsf{\textbf{Non-goals.}}
Label noise may also come from incorrect mention boundaries and wrong mapping of mentions to KB entities.
This work relies on existing entity linking tools~\cite{shen2014entity} to provide decent entity mention detection and resolution results (\eg, leftmost column of Fig.~\ref{figure:framework_overview}), but we do not address their limits here.
We also assume human-curated target type hierarchies are given for the task (It is out of the scope of this study to generate the type hierarchy \begin{small}$\Y$\end{small}).

\section{Label Noise Reduction}
\label{sec:method}
This section lays out the framework. As the candidate type sets in the training corpus contain ``false" types, \textit{supervised} learning techniques (\eg, multi-label learning~\cite{ling2012fine}, hierarchical classification~\cite{yosef2012hyena}) may generate predictions biased to the incorrect type labels~\cite{gillick2014context}.
Our solution casts the problem as a \textit{weakly-supervised} learning task, which aims to derive the relatedness between mentions and their candidate types using both corpus-level statistics and KB facts.

Specifically, each entity type is treated as an \textit{individual object} to be modeled. As type assignment on each mention is \textit{noisy}, we adopt ideas from partial label learning~\cite{cour2011learning} to carefully model mention-type associations, and extract a set of text features for each mention to assist in modeling its true types. In order to capture the semantic similarity between types, we further derive type correlation from two different sources, \ie, KB and the given type hierarchy.

\medskip
\noindent \textsf{\textbf{Framework Overview.}}
We propose a \textit{graph-based partial-label embedding framework} (see also Fig.~\ref{figure:framework_overview}) as follows:
\begin{enumerate}[leftmargin=13pt]\itemsep+0.2cm
\item Generate text features for each entity mention \begin{small}$m_i\in\M$\end{small}, and construct a heterogeneous graph using three kinds of objects in the corpus, namely entity mentions \begin{small}$\M$\end{small}, target types \begin{small}$\Y$\end{small} and text features (denoted as \begin{small}$\F$\end{small}), to encode aforementioned signals in a unified form (Sec.~\ref{subsec:graph}).

\item Perform joint embedding of the constructed graph $G$ into the same low-dimensional space where, in that space, close objects (\ie, whose embedding vectors have high similarity score) tend to also share the same types (Sec.~\ref{subsec:model}).

\item For each mention \begin{small}$m_i$\end{small} (in set \begin{small}$\M$\end{small}), search its candidate type sub-tree \begin{small}$\Y_i$\end{small} in a top-down manner and estimate the true type-path \begin{small}$\Y_i^*$\end{small} from learned embeddings (Sec.~\ref{subsec:algorithm}).
\end{enumerate}

\begin{table*}[t]
\vspace{-0.0cm}
\begin{small}
\begin{center}
\vspace{0.0cm}
\begin{tabularx}{\textwidth}{>{\hsize=0.5\hsize}X|>{\hsize=1.5\hsize}X| X}
\hline
\textbf{Feature} & \textbf{Description} & \textbf{Example} \\
\hline
Head & Syntactic head token of the mention & ``HEAD\_Turing"\\ 
Token & Tokens in the mention & ``Turing", ``Machine" \\ 
POS & Part-of-Speech tag of tokens in the mention & ``NN" \\ 
Character & All character trigrams in the head of the mention & ``:tu", ``tur", ..., ``ng:" \\ 
Word Shape & Word shape of the tokens in the mention & ``Aa" for ``Turing"\\ 
Length & Number of tokens in the mention & ``2" \\ 
Context & Unigrams/bigrams before and after the mention & ``CXT\_B:Maserati ,",~``CXT\_A:and the" \\ 
Brown Cluster & Brown cluster ID for the head token (learned using $\D$) & ``4\_1100",~``8\_1101111",~``12\_111011111111" \\ 
Dependency & Stanford syntactic dependency~\cite{manning2014stanford} associated with the head token & ``GOV:nn", ``GOV:turing" \\ 
\hline
\end{tabularx}
\vspace{-0.2cm}
\caption{Text features used in this paper. ``\textit{Turing Machine}" is used as an example mention from ``\textit{The band’s former drummer Jerry Fuchs---who was also a member of Maserati, \textbf{Turing Machine} and The Juan MacLean---died after falling down an elevator shaft.}".}
\label{table:features}
\vspace{-0.2cm}
\end{center}
\end{small}
\end{table*}

\subsection{Construction of Graphs}
\label{subsec:graph}
To capture the shallow syntax and distributional semantics of a mention \begin{small}$m_i\in\M$\end{small}, we extract various features from both \begin{small}$m_i$\end{small} itself (\eg, head token) and its context \begin{small}$c_i$\end{small} (\eg, bigram).
Table~\ref{table:features} lists the set of text features used in this work, which is similar to those used in~\cite{Yogatama2015embedding,ling2012fine}.
We denote the set of \begin{small}$M$\end{small} unique features of \begin{small}$\M$\end{small} extracted from \begin{small}$\D$\end{small} as \begin{small}$\F = \{f_j\}_{j=1}^M$\end{small}.
Details of feature generation are introduced in Sec.~\ref{subsec:data_preparation}.

With entity mentions \begin{small}$\M$\end{small}, text features \begin{small}$\F$\end{small} and target types \begin{small}$\Y$\end{small}, we build a heterogeneous graph \begin{small}$G$\end{small} to unify three kinds of links: \textit{mention-type link} represents each mention's candidate type assignment; \textit{mention-feature link} captures corpus-level co-occurrences between mentions and text features; and \textit{type-type link} encodes the type correlation derived from KB or target type hierarchy.
This leads to three subgraphs \begin{small}$G_{MY}$\end{small}, \begin{small}$G_{MF}$\end{small}, and \begin{small}$G_{YY}$\end{small}, respectively. 
\nop{We introduce the construction of them in the rest of the section.}

\medskip
\noindent \textsf{\textbf{Mention-Type Association Subgraph.}}
In the automatically labeled training corpus \begin{small}$\D = \big\{(m_i, c_i, \Y_i)\big\}$\end{small}, each mention \begin{small}$m_i$\end{small} is assigned a set of candidate types \begin{small}$\Y_i$\end{small} from the target type set \begin{small}$\Y$\end{small}. This naturally forms a bipartite graph between entity mentions \begin{small}$\M$\end{small} and target types \begin{small}$\Y$\end{small}, where each mention \begin{small}$m_i\in\M$\end{small} is linked to its candidate types \begin{small}$\Y_i$\end{small} with binary weight, \ie, \begin{small}$G_{MY} = \big\{(m_i, y_k)~|~y_k\in\Y_i,~m_i\in\M\big\}$\end{small}; \begin{small}$w_{ik}=1$\end{small} if \begin{small}$(m_i, y_k)\in G_{MY}$\end{small} and \begin{small}$w_{ik}=0$\end{small} otherwise.

Existing embedding methods rely on either the \textit{local consistent assumption}~\cite{he2004locality} (\ie, objects strongly connected tend to be similar), or the \textit{distributional assumption}~\cite{mikolov2013distributed} (\ie, objects sharing similar neighbors tend to be similar) to model graph structures. However, some links are ``false" links in the constructed mention-type subgraph---adopting the above assumptions may incorrectly yield mentions of different types having similar embeddings. For example, in Fig.~\ref{figure:framework_overview}, ``\textit{Hillary Clinton}" in S1 and ``\textit{Trump}" in S3 have several candidate types in common (thus high distributional similarity), but their true types are different (\ie, \texttt{politician} versus \texttt{businessman}).
Instead of defining a binary variable to indicate whether a mention-type link is true or not, we specify the likelihood of a mention-type link being true as the \textit{relevance} between the corresponding mention and type, and progressively estimate the relevance by incorporating other \textit{side signals} (\eg, text features, type correlation).
We propose to model mention-type links based on the following hypothesis.
\begin{hypothesis}[Partial Label Association]
\label{hypothesis:partial_label}
A mention should be embedded closer to its most relevant candidate type than to any other non-candidate type, yielding higher similarity between the corresponding embedding vectors.
\end{hypothesis}

During model learning, relevance between an entity mention and its candidate type is measured by the similarity between their current estimated embeddings. Text features, as complements to mention-candidate type links, also participate in modeling the mention embeddings, and help identify a mention's most relevant type. In sentence S1 of Fig.~\ref{figure:framework_overview}, context words \textit{democratic} and \textit{presidential} infer that type \texttt{politician} is more relevant than type \texttt{actor} for mention ``\textit{Hillary Clinton}".
This hypothesis assumes that the embeddings of two mentions will be close if and only if their \textit{most relevant candidate types} are similar.


\medskip
\noindent \textsf{\textbf{Mention-Feature Co-occurrence Subgraph.}}
Intuitively, entity mentions sharing many text features (\ie, with similar distributions over \begin{small}$\F$\end{small}) tend to have close type semantics; and text features which co-occur with many entity mentions in the corpus (\ie, with similar distributions over \begin{small}$\M$\end{small}) likely represent similar entity types.
The following hypothesis guides our modeling of mention-text feature co-occurrences.
\begin{hypothesis}[Mention-Feature Co-occurrences]
\label{hypothesis:mention_feature_cooccurrence}
If two entity mentions share similar features, they should be close to each other in the embedding space (\ie, high similarity score).  If two features co-occur with a similar set of mentions, their embedding vectors tend to be similar.
\end{hypothesis}
In Fig.~\ref{figure:framework_overview}, for example, mentions ``\textit{Donald Trump}" in S2 and ``\textit{Trump}" in S4 share multiple features (\eg, \textit{Trump}, \textit{presidential} and \textit{campaign}), and thus are likely of the same type \texttt{politician}. Conversely, features \textit{campaign} and \textit{presidential} likely represent the same type \texttt{politician} since they co-occur with similar sets of mentions in the corpus.

Formally, we form binary links between mentions and their text features to construct a mention-feature co-occurrence subgraph, \ie, \begin{small}$w_{ij}=1$\end{small} if feature \begin{small}$f_j\in\F$\end{small} is extracted for mention \begin{small}$m_i\in\M$\end{small}; and \begin{small}$w_{ij}=0$\end{small} otherwise. We use \begin{small}$G_{MF} = \big\{(m_i, f_j)~|~w_{ij}=1,~m_i\in\M,~f_j\in\F\big\}$\end{small} to denote the subgraph.

\begin{figure}
\centering
\begin{small}
\vspace{-0.0cm}
\includegraphics[width = 87 mm]{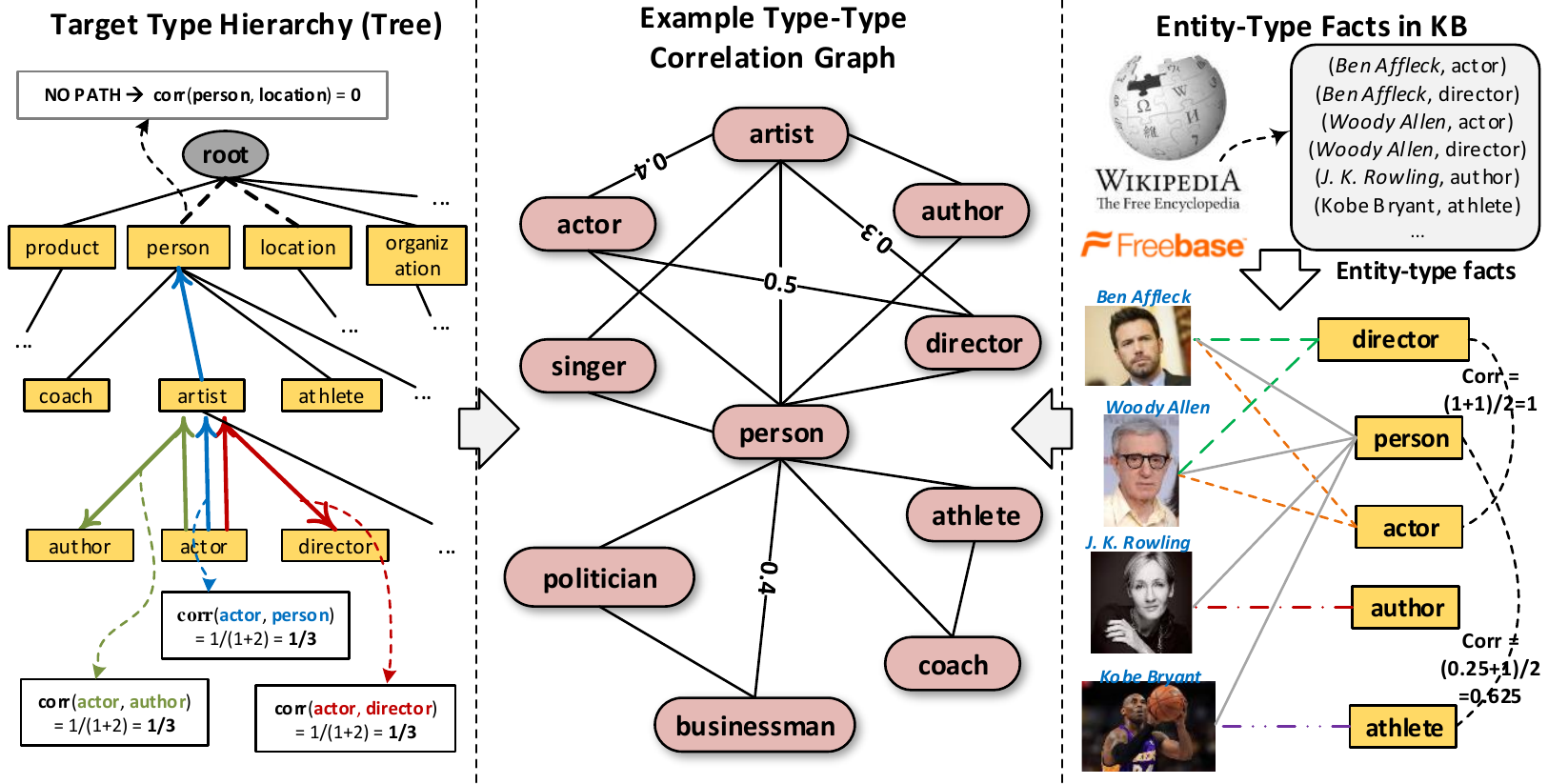}
\vspace{-0.5cm}
\caption{Example of constructing type correlation graph.}
\label{figure:type_correlation_graph}
\end{small}
\vspace{-0.0cm}
\end{figure}

\medskip
\noindent \textsf{\textbf{Type Correlation Subgraphs.}}
In target type hierarchy \begin{small}$\Y$\end{small}, types closer to each other (\ie, shorter path) tend to be more related (\eg, \texttt{actor} is more related to \texttt{artist} than to \texttt{person} in the left column of Fig.~\ref{figure:type_correlation_graph}).
In KB \begin{small}$\Psi$\end{small}, types assigned to similar sets of entities should be more related to each other than those assigned to quite different entities~\cite{Jiang2015typeHierarchy} (\eg, \texttt{actor} is more related to \texttt{director} than to \texttt{author} in the right column of Fig.~\ref{figure:type_correlation_graph}).
We propose to model type correlation based on the following hypothesis.
\begin{hypothesis}[Type Correlation]
\label{hypothesis:type_correlation}
If high correlation exists between two target types based on either type hierarchy or KB, they should be embedded close to each other.
\end{hypothesis}

We build a homogeneous graph \begin{small}$G_{YY}$\end{small} to represent the correlation between types.
A simple way to measure correlation between two types is to use their distance in the target type hierarchy (tree). Specifically, a link \begin{small}$(y_k, y_{k'})$\end{small} is formed if there exists a path between types \begin{small}$y_k$\end{small} and \begin{small}$y_{k'}$\end{small} in \begin{small}$\Y$\end{small} (paths passing root node are excluded).
We define the weight of link \begin{small}$(y_k, y_{k'})\in G_{YY}$\end{small} as \begin{small}$w_{kk'} = 1 / \big(1+ \rho(y_k, y_{k'})\big)$\end{small}, where \begin{small}$\rho(y_k, y_{k'})$\end{small} denotes the length of the shortest path between types \begin{small}$y_k$\end{small} and \begin{small}$y_{k'}$\end{small} in \begin{small}$\Y$\end{small}. Although using shortest path to compute type correlation is efficient, its accuracy is limited---It is not always true that a type (\eg, \texttt{athlete}) is more related to its parent type (\ie, \texttt{person}) than to its sibling types (\eg, \texttt{coach}), or that all sibling types are equally related to each other (\eg, \texttt{actor} is more related to \texttt{director} than to \texttt{author}).

An alternative approach to avoid this accuracy issue is to exploit entity-type facts \begin{small}$\T_\Psi$\end{small} in KB to measure type correlation.
Given two target types \begin{small}$y_k, y_{k'}\in\Y$\end{small}, the correlation between them is proportional to the number of entities they share in the KB.
Let \begin{small}$\E_k$\end{small} denote the set of entities assigned with type \begin{small}$y_k$\end{small} in KB, \ie, \begin{small}$\E_k = \big\{e~|~(e,y_k)\in\T_\Psi\big\}$\end{small}. The weight \begin{small}$w_{kk'}$\end{small} of link \begin{small}$(y_k, y_{k'})\in G_{YY}$\end{small} is defined as follows.
\begin{align}
\label{eq:KB_type_correlation}
w_{kk'} = \Big(\big|\E_k\cap\E_{k'}\big|/\big|\E_k\big| + \big|\E_k\cap\E_{k'}\big|/\big|\E_{k'}\big|\Big)/2,
\end{align}
where \begin{small}$|\E_k|$\end{small} denotes the size of set \begin{small}$\E_k$\end{small}.
We compare these two methods for measuring type correlation 
in our experiments.
Entity-entity facts of various relationships in the KB can also be utilized to model type correlation, as discussed in KB embedding~\cite{hu2015hiearchy,bordes2013translating}. We leave this as future work.

\subsection{Heterogeneous Partial-Label Embedding}
\label{subsec:model}
This section follows notations in Table~\ref{table:notation} to formulate a joint optimization problem for embedding the constructed heterogeneous graph \begin{small}$G$\end{small} into a $d$-dimensional vector space.

A straightforward solution is to model the whole graph with the local consistency objective~\cite{he2004locality}. Such a solution encounters several problems: False candidate types negatively impact the ability of the model to determine mention's true types, and the mention-feature links are too sparse to model mention's types. As such, the learned embeddings may not accurately capture relatedness between mentions and types.

In our solution, we formulate a novel optimization objective, by extending a margin-based rank loss to model noisy mention-type links (\ie, \begin{small}$G_{MY}$\end{small}) and leveraging the distributional assumption~\cite{mikolov2013distributed} to model subgraphs \begin{small}$G_{MF}$\end{small} and \begin{small}$G_{YY}$\end{small}.

\medskip
\noindent \textsf{\textbf{Modeling Mention-Type Association.}}
To effectively model the \textit{noisy} mention-type links in subgraph \begin{small}$G_{MY}$\end{small}, we extend the margin-based loss in~\cite{nguyen2008classification} (used to learn linear classifiers) to enforce Hypothesis~\ref{hypothesis:partial_label}. The intuition of the loss is simple: for mention \begin{small}$m_i$\end{small}, the maximum score associated with its candidate types \begin{small}$\Y_i$\end{small} is greater than the maximum score associated with any other non-candidate types \begin{small}$\Yn_i=\Y~\backslash~\Y_i$\end{small}, where the scores are measured using current embedding vectors.

Specifically, we use vectors $\bu_i,~\bv_k\in\R^d$ to represent mention \begin{small}$m_i\in\M$\end{small} and type \begin{small}$y_k\in\Y$\end{small} in the $d$-dimensional embedding space, respectively. The score of \begin{small}$(m_i, y_k)$\end{small} is defined as the dot product of their embeddings, \ie, \begin{small}$\s (m_i, y_k) = \bv_k^T\bu_i$\end{small}. We define the \textit{partial-label loss}  \begin{small}$\ell_i$\end{small} for \begin{small}$m_i\in\M$\end{small} as follows.
\begin{align}
\label{eq:mention_type_loss}
\ell_{i} = \max\Big\{0, 1 - \Big[\max_{y\in\Y_i}s(m_i, y)  - \max_{y'\in \Yn_i}s(m_i, y')\Big]\Big\}.
\end{align}

Minimizing \begin{small}$\ell_i$\end{small} encourages a \textit{large margin} between the maximum scores \begin{small}$\max_{y\in\Y_i}s(m_i, y)$\end{small} and \begin{small}$\max_{y'\in \Yn_i}s(m_i, y')$\end{small}. This forces \begin{small}$m_i$\end{small} to be embedded closer to the most ``relevant" type in the noisy candidate type set, \ie, \begin{small}$y^*=\argmax_{y\in\Y_i}s(m_i,y)$\end{small}, than to \textit{any other} non-candidate types (\ie, Hypothesis~\ref{hypothesis:partial_label}).
This constrasts sharply with multi-label learning~\cite{yosef2012hyena}, where a large margin is enforced between \textit{all} candidate types and non-candidate types without considering noisy types.

\begin{table}[t]
\vspace{-0.0cm}
\begin{scriptsize}
\begin{center}
\begin{tabularx}{\linewidth}{l|X}
\hline
$\D$ & Automatically generated training corpus \\ 
$\M=\{m_i\}_{i=1}^N$ & Entity mentions in $\D$ (size $N$) \\ 
$\Y=\{y_k\}_{k=1}^K$ & Target entity types (size $K$)\\ 
$\Y_i$ & Candidate types of $m_i$\\
$\Yn_i$ & Non-candidate types of $m_i$, \ie, $\Yn_i=\Y~\backslash~\Y_i$ \\
$\F=\{f_j\}_{j=1}^M$ & Text features in $\D$ (size $M$)\\ 
$\bu_i \in\R^d$ & Embedding of mention $m_i$ (dim. $d$)\\ 
$\bc_j\in\R^d$ & Embedding of feature $f_j$ (dim. $d$)\\ 
$\bv_k, \bv'_k\in\R^d$ & Embeddings of type $y_k$ on two views (dim. $d$)\\
\hline
\end{tabularx}
\vspace{-0.3cm}
\caption{Notations.}
\label{table:notation}
\vspace{-0.2cm}
\end{center}
\end{scriptsize}
\end{table}

\medskip
\noindent \textsf{\textbf{Modeling Mention-Feature Co-occurrences.}}
Hypothesis~\ref{hypothesis:mention_feature_cooccurrence} models mention-feature links based on the idea that \textit{nodes with similar distributions over neighbors are similar to each other}. This idea is similar to that found in Second-order Proximity model~\cite{tang2015line}, and Skip-gram model~\cite{mikolov2013distributed}---it models text corpora following the distributional hypothesis~\cite{harris1954distributional} which says that you should know a word by the company it keeps.

Formally, we introduce vector \begin{small}$\bc_j\in\R^d$\end{small} to represent feature \begin{small}$f_j\in\F$\end{small} in the embedding space. Following second-order proximity~\cite{tang2015line}, we define the probability of feature \begin{small}$f_j$\end{small} generated by mention \begin{small}$m_i$\end{small} for each link \begin{small}$(m_i, f_j)\in G_{MF}$\end{small} as follows.
\begin{footnotesize}
\begin{align}
\label{eq:mention_feature_cond_prob}
p(f_j|m_i) = \e(\bc_j^T\bu_i)\big/\sum_{f_{j'}\in\F}\e(\bc_{j'}^T\bu_i).
\end{align}
\end{footnotesize}

High conditional probability \begin{small}$p(f_j|m_i)$\end{small} indicates that embeddings of \begin{small}$m_i$\end{small} and \begin{small}$f_j$\end{small} are similar. Following Hypothesis~\ref{hypothesis:mention_feature_cooccurrence}, we enforce the conditional distribution specified by embeddings, \ie, \begin{small}$p(\cdot|m_i)$\end{small} to be close to the empirical distribution (\ie, link distribution of \begin{small}$m_i$\end{small} to \begin{small}$\F$\end{small} in subgraph \begin{small}$G_{MF}$\end{small}), which can be achieved by minimizing the following objective~\cite{tang2015line}.
\begin{align}
\label{eq:mention_feature_obj}
\O_{MF} = -\sum_{(m_i,f_j)\in G_{MF}}w_{ij}\cdot\log~p(f_j|m_i).
\end{align}

Optimizing \begin{small}$\O_{MF}$\end{small} with \begin{small}$p(f_j|m_i)$\end{small} defined by Eq.~\eqref{eq:mention_feature_cond_prob} is computationally expensive since it involves summation over all the features. We adopt the negative sampling method~\cite{mikolov2013distributed} to sample multiple \textit{negative} features for each link $(m_i,f_j)$, according to some \textit{noise distribution}. The method replaces \begin{small}$\log~p(f_j|m_i)$\end{small} in Eq.~\eqref{eq:mention_feature_obj} with the following function.
\begin{align}
\label{eq:mention_feature_neg_sampling}
\log~\sigma(\bc_j^T\bu_i) + \sum_{z=1}^Z\EE_{f_l\sim P_n(f)}\big[\log~\sigma(-\bc_l^T\bu_i)\big],
\end{align}
where \begin{small}$\sigma(x)=1/\big(1+\exp(-x)\big)$\end{small} is the sigmoid function. The first term in Eq.~\eqref{eq:mention_feature_neg_sampling} models the observed links in \begin{small}$G_{MF}$\end{small}, and the second term models the \begin{small}$Z$\end{small} negative features sampled from the noise distribution \begin{small}$P_n(f) \propto D_f^{3/4}$\end{small} over all features \begin{small}$\F$\end{small}~\cite{mikolov2013distributed}. Here \begin{small}$D_f$\end{small} denotes the degree of feature $f$ in \begin{small}$G_{MF}$\end{small}.

\medskip
\noindent \textsf{\textbf{Modeling Type Correlation.}}
Type correlation links can be modeled with a method similar to that used in modeling the mention-feature subgraph---two types are similar to each other if they are correlated to the same set of types (\ie, Hypothesis~\ref{hypothesis:type_correlation}).
As link \begin{small}$(m_i,f_j)$\end{small} in bipartite graph \begin{small}$G_{MF}$\end{small} is \textit{directed}, we treat each \textit{undirected} link \begin{small}$(y_k,y_{k'})$\end{small} in the homogeneous graph \begin{small}$G_{YY}$\end{small} as two directed links~\cite{tang2015pte}. Hypothesis~\ref{hypothesis:type_correlation} can be modeled by minimizing the following objective.
\begin{align*}
\O_{YY} = -\sum_{(y_k,y_{k'})\in G_{YY}}w_{kk'}\Big[\log~p(y_{k'}|y_{k})+\log~p(y_k|y_{k'})\Big].
\end{align*}
This enforces the conditional distributions specified by embeddings to be close to its empirical distributions in terms of \textit{both directions} of the link \begin{small}$(y_k,y_{k'})$\end{small}. We use two vectors \begin{small}$\bv_k,~\bv'_k\in\R^d$\end{small} to represent each type \begin{small}$y_k\in\Y$\end{small} in the embedding space, where \begin{small}$\bv'_k$\end{small} serves as the ``context" view of \begin{small}$y_k$\end{small}~\cite{tang2015line}. Following a similar negative sampling procedure as that in Eq.~\eqref{eq:mention_feature_neg_sampling}, we define  \begin{small}$\log~p(y_{k'}|y_{k})$\end{small} as follows.
\begin{align}
\label{eq:type_neg_sampling}
\log~\sigma({\bv'_k}^{T}\bv_k) + \sum_{z=1}^Z\EE_{y_l\sim P_n(y)}\big[\log~\sigma(-{\bv'_l}^{T}\bv_k)\big].
\end{align}
Similar to the derivation of \begin{small}$\log~p(y_{k'}|y_{k})$\end{small} in Eq.~\eqref{eq:type_neg_sampling}, we can define the log probability \begin{small}$\log~p(y_{k}|y_{k'})$\end{small}.

\begin{table}[t]
\vspace{-0.0cm}
\begin{scriptsize}
\begin{center}
\begin{tabularx}{\linewidth}{|l|x|}
\hline
\textbf{Similar types}&\textbf{Similar features} \\ \hline
\texttt{organization stock\_exchange} & CXT\_A:Trans\_World \\
\texttt{organization government} & CXT\_A:Automobile\_Insurance \\
\texttt{organization education} & CXT\_A:dual\_trading\\
\hline
\end{tabularx}
\vspace{-0.3cm}
\caption{Example similar types and features for feature ``CXT\_B:Deutsche\_Bank'' based on the learned PLE embeddings.}
\label{table:feature_example}
\vspace{-0.2cm}
\end{center}
\end{scriptsize}
\end{table}

\medskip
\noindent \textsf{\textbf{The Joint Optimization Problem.}}
Our goal is to embed the heterogeneous graph \begin{small}$G$\end{small} into a $d$-dimensional vector space, following the three proposed hypotheses in Sec.~\ref{subsec:graph}. Intuitively, one can \textit{collectively} minimize the objectives of the three subgraphs \begin{small}$G_{MY}$\end{small}, \begin{small}$G_{MF}$\end{small} and \begin{small}$G_{YY}$\end{small}, as mentions \begin{small}$\M$\end{small} and types \begin{small}$\Y$\end{small} are shared across them. To achieve the goal, we formulate a joint optimization problem as follows.
\begin{align}
\label{eq:objective}
\min_{\{\bu_i\}_{i=1}^N,\{\bc_j\}_{j=1}^M,\{\bv_k,\bv'_k\}_{k=1}^K} \O = \O_{MY} + \O_{MF} + \O_{YY},
\end{align}
where objective \begin{small}$\O_{MY}$\end{small} of the subgraph \begin{small}$G_{MY}$\end{small} is specified by aggregating the partial-label loss defined in Eq.~\eqref{eq:mention_type_loss} across all the mentions \begin{small}$\M$\end{small}, along with $\ell_2$-regularizations on $\{u_i\}_{i=1}^N$ and $\{v_k\}_{k=1}^K$ to control the scale of the embeddings~\cite{nguyen2008classification}.
\begin{align}
\label{eq:mention_type_obj}
\O_{MY} = \sum_{i=1}^N \ell_i + \frac{\lambda}{2} \sum_{i=1}^N\|\bu_i\|_2^2 + \frac{\lambda}{2} \sum_{k=1}^K \|\bv_k\|_2^2.
\end{align}
Tuning parameter \begin{small}$\lambda>0$\end{small} is used to control the amount of regularization on the embeddings.
In Eq.~\eqref{eq:objective}, one can also minimize the weighted combination of the three subgraph objectives to model the importance of different signals, where weights could be  manually determined or automatically learned from data. We leave this as future work.
By solving the optimization problem in Eq.~\eqref{eq:objective}, we are able to represent every node in \begin{small}$G$\end{small} with a $d$-dimensional vector.

\subsection{Model Learning and Inference}
\label{subsec:algorithm}
We propose an alternative minimization algorithm based on block-wise coordinate descent schema~\cite{tseng2001convergence} to \textit{jointly} optimize the objective \begin{small}$\O$\end{small} in Eq.~\eqref{eq:objective}. In each iteration, the algorithm goes through links in \begin{small}$G$\end{small} to sample negative links, and update each embedding based on the derivatives.

\begin{algorithm}[t]
\begin{small}
\DontPrintSemicolon
\KwIn{$G=\{G_{MY}, G_{MF}, G_{YY}\}$, regularization parameter $\lambda$, learning rate $\alpha$, number of negative samples $Z$}
\KwOut{entity mention embeddings $\{\bu_i\}_{i=1}^N$, feature embeddings $\{\bc_j\}_{j=1}^M$, type embeddings $\{\bv_k\}_{k=1}^K$}
Initialize: $\{\bu_i\}$, $\{\bc_j\}$, and $\{\bv_k\}$ as random vectors\;
\While{$\O$ in Eq.~\eqref{eq:objective} not converge}{
	\For{each link in $G_{MF}$ and $G_{YY}$}{
      Draw $Z$ negative links from noise distribution $P_n(\cdot)$
    	}	
	\For{$m_i\in\M$}{
      $\bu_i\gets\bu_i - \alpha\cdot\partial\O/\partial\bu_i$  with $\partial\O/\partial\bu_i$ defined in Eq.~\eqref{eq:derivative_u}\;
    	}
    	\For{$f_j\in\F$}{
      $\bc_j\gets\bc_j - \alpha\cdot\partial\O/\partial\bc_j$ using $\partial\O/\partial\bc_j$ defined in Eq.~\eqref{eq:derivative_c}\;
    	}
    	\For{$y_k\in\Y$}{
      $\bv_k\gets\bv_k - \alpha\cdot\partial\O/\partial\bv_k$ based on $\partial\O/\partial\bv_k$ in Eq.~\eqref{eq:derivative_v}\;
      $\bv'_k\gets\bv'_k - \alpha\cdot\partial\O/\partial\bv'_k$ using $\partial\O/\partial\bv'_k$ in Eq.~\eqref{eq:derivative_vp}\;

    	}
}
\caption{{\small Model Learning of PLE}}
\label{algorithm:PLE}
\end{small}
\end{algorithm}

We first take the derivative of \begin{small}$\O$\end{small} with respect to \begin{small}$\{\bu_i\}$\end{small} while fixing other variables. 
A similar procedure to that in~\cite{nguyen2008classification}  is followed to calculate the derivative for partial-label loss.
\begin{small}
\begin{align}
\label{eq:derivative_u}
\frac{\partial \O}{\partial \bu_i} =& \lambda\bu_i + \one\Big(\max_{y_k\in\Y_i}\bu_i^T\bv_k - \max_{y_{k'}\in\Yn_i}\bu_i^T\bv_{k'} < 1\Big)(\bv_i^+ - \bv_i^-) \nonumber\\
& -\sum_{f_j\in\F_i}\Big[\sigma (-\bc_j^T\bu_i)\bc_j - \sum_{z=1}^Z\EE_{f_l\sim P_n(f)}[\sigma (\bc_l^T\bu_i)\bc_l]\Big],
\end{align}
\end{small}
where \begin{small}$\one(\cdot)$\end{small} denotes the indicator function, and \begin{small}$\F_i=\{f~|~(m_i,f)\\\in G_{MF}\}$\end{small} denotes features linked to \begin{small}$m_i$\end{small} in \begin{small}$G_{MY}$\end{small}. 
We use \begin{small}$\bv_i^+ = \argmax_{y_l\in\Y_i}~\bu_i^T\bv_l$ and $\bv_i^- = \argmax_{y_l\in\Yn_i}~\bu_i^T\bv_l$\end{small} to denote the embeddings of the most relevant types in \begin{small}$m_i$\end{small}'s candidate type set \begin{small}$\Y_i$\end{small} and non-candidate set \begin{small}$\Yn_i$\end{small}, respectively.

The first two terms in Eq.~\eqref{eq:derivative_u} adjust \begin{small}$\bu_i$\end{small} to ensure sufficient difference (margin) exists between its similarity to the most relevant candidate type and that to any non-candidate type.
The last part requires \begin{small}$\bu_i$\end{small} to be close to (different from) its linked (unlinked) features in \begin{small}$G_{MF}$\end{small}, respectively.

Second, we fix \begin{small}$\{\bu_i\}$\end{small} and \begin{small}$\{\bv_k\}$\end{small} to compute the derivative of \begin{small}$O$\end{small} with respect to \begin{small}$\{\bc_j\}$\end{small}. Let \begin{small}$\M_j=\{m~|~(m,f_j)\in G_{MF}\}$\end{small} denote the mentions linked to feature \begin{small}$f_j$\end{small} in graph \begin{small}$G_{MF}$\end{small}.
\begin{small}
\begin{align}
\label{eq:derivative_c}
\frac{\partial \O}{\partial \bc_j} = 
-\sum_{m_i\in\M_j}\sigma (-\bc_j^T\bu_i)\bu_i 
+ \sum_{z=1}^{Z\cdot|G_{MF}|}\EE_{f_l\sim P_n(f)}^{~l=j}\Big[\sigma (\bc_j^T\bu)\bu\Big].
\end{align}
\end{small}
The first part in Eq.~\eqref{eq:derivative_c} models the observed links between feature \begin{small}$f_j$\end{small} and other mentions in graph \begin{small}$G_{MF}$\end{small}. The second part models negative samples drawn from links in \begin{small}$G_{MF}$\end{small} (\ie, with size \begin{small}$Z|G_{MF}|$\end{small}) which involve feature \begin{small}$f_j$\end{small}. We use \begin{small}$\EE_{f_l\sim P_n(f)}^{l=j}[\cdot]$\end{small} to denote the negative sampling process.

Finally, we compute the derivatives for \begin{small}$\{\bv_k,\bv'_k\}$\end{small} by fixing other variables. We use \begin{small}$\N_k=\{y~|~(y,y_k)\in G_{YY}\}$\end{small} to denote the set of types linked to type \begin{small}$y_k$\end{small} in graph \begin{small}$G_{YY}$\end{small}.
\begin{small}
\begin{align}
\label{eq:derivative_v}
\frac{\partial \O}{\partial \bv_k} = \lambda\bv_k 
& + \sum_{i=1}^N\one\Big(\max_{y_l\in\Y_i}\bu_i^T\bv_l - \max_{y_{l'}\in\Yn_i}\bu_i^T\bv_{l'} < 1\Big)\cdot \bu_i^k \\ 
- \sum_{y_{k'}\in \N_k}& w_{kk'}\Big[\sigma (-{\bv'}_{k'}^T\bv_k)\bv'_{k'}-\sum_{z=1}^Z\EE_{y_l\sim P_n(y)}\big[\sigma ({\bv'}_l^T\bv_k)\bv'_l\big]\Big],
\nonumber
\end{align}
\end{small}
where for each $k$ the vector \begin{small}$\bu_i^k$\end{small} is defined as follow.
\begin{footnotesize}
\begin{align*}
\bu_i^k = &\Big[\one\big(y_k=\argmax\limits_{y_{l}\in\Yn_i}~\bu_i^T\bv_{l}\big) - \one\big(y_k=\argmax\limits_{y_l\in\Y_i}~\bu_i^T\bv_l\big)\Big]\bu_i.
\end{align*}
\end{footnotesize}

The derivative with respect to \begin{small}$\{\bv'_k\}$\end{small} can be computed in a way similar to Eq.~\eqref{eq:derivative_c}, which models both the observed links in \begin{small}$G_{YY}$\end{small} and the negative samples of the observed links.
\begin{small}
\begin{align}
\label{eq:derivative_vp}
\frac{\partial \O}{\partial \bv'_k} = 
-\sum_{y_{k'}\in \N_k}\sigma (-{\bv'}_k^T\bv_{k'})\bv_{k'}
+ \sum_{z=1}^{Z\cdot|G_{YY}|}\EE_{y_l\sim P_n(y)}^{~l=k}\Big[\sigma ({\bv'}_k^T\bv)\bv\Big].
\end{align}
\end{small}

Algorithm~\ref{algorithm:PLE} summarizes our algorithm. Eq.~\eqref{eq:objective} can also be solved by a mini-batch extension of the Pegasos algorithm~\cite{shalev2011pegasos}, which is a stochastic sub-gradient descent method and thus can efficiently handle massive text corpora. Due to lack of space, we do not include derivation details here.

\begin{algorithm}[t]
\begin{small}
\DontPrintSemicolon
\KwIn{candidate type sub-tree $\{\Y_i\}$, mention embeddings $\{\bu_i\}$, type embeddings $\{\bv_k\}$, threshold $\eta$}
\KwOut{estimated type-path $\{\Y^*_i\}$ for $m_i\in\M$}
\For{$m_i\in\M$}{
	Initialize: $\Y_i^*$ as $\emptyset$, $r$ as the root of $\Y$\;
	\While{$\C_i(r)\neq\emptyset$}{
		$r\gets\argmax_{y_k\in\C_i(r)}s(\bu_i, \bv_k)$\;
		\eIf{$s(\bu_i,\bv_r) > \eta$}{
			Update the type-path: $\Y^*_i\gets\Y^*_i\bigcup\{r\}$\;
		}
		{
		\Return{$\Y^*_i$} as the estimated type-path for $m_i$\;
		}
    	}
}
\caption{{\small Type Inference}}
\label{algorithm:type_inference}
\end{small}
\end{algorithm}

\medskip
\noindent \textsf{\textbf{Type Inference.}}
With the learned mention embeddings \begin{small}
$\{\bu_i\}$\end{small} and type embeddings \begin{small}
$\{\bv_k\}$\end{small}, we perform top-down search in the candidate type sub-tree \begin{small}
$\Y_i$\end{small} to estimate the correct type-path \begin{small}
$\Y^*_i$\end{small}. Starting from the tree's root (denoted as $r$), we recursively find the best type among the children types (denoted as \begin{small}$\C_i(r)$\end{small}) by measuring the dot product of the corresponding mention and type embeddings, \ie, \begin{small}$s(\bu_i,\bv_k)$\end{small}. The search process stops when we reach to leaf type, or the similarity score is below a pre-defined threshold $\eta>0$. Algorithm~\ref{algorithm:type_inference} summarizes the proposed type inference process.

\medskip
\noindent \textsf{\textbf{Computational Complexity Analysis.}}
In graph construction, the cost of building subgraph \begin{small}$G_{YY}$\end{small} is \begin{small}$O(K^2I)$\end{small}, where \begin{small}$I$\end{small} is the average number of entities associated with a type in the KB. Building \begin{small}$G_{MY}$\end{small} and \begin{small}$G_{MF}$\end{small} takes \begin{small}$O(N)$\end{small} time.

Let \begin{small}$E$\end{small} be the total number of links in \begin{small}$G$\end{small}.
By alias table method~\cite{tang2015line}, sampling a negative link takes constant time and setting up alias tables takes \begin{small}$O(N+M+K)$\end{small} time for all the nodes in \begin{small}$G$\end{small}. In each iteration of Algorithm~\ref{algorithm:PLE}, optimization with negative sampling and partial labels takes \begin{small}$O\big(d(Z+K)E\big)$\end{small} time. Supposing the algorithm stops after \begin{small}$T$\end{small} iterations (\begin{small}$T<50$\end{small} in our experiments), the overall time complexity of PLE is \begin{small}$O\big(dT(Z+K)E\big)$\end{small}, which is linear to the number of links \begin{small}$E$\end{small} and does not depend on the number of nodes in \begin{small}$G$\end{small}.
\section{Experiments}
\label{sec:experiments}
\subsection{Data Preparation and Experiment Setting}
\label{subsec:data_preparation}
Our experiments use three public datasets\footnote{\small Codes and datasets used in this paper can be downloaded at: \url{https://github.com/shanzhenren/PLE}.}. (1) \textbf{Wiki}~\cite{ling2012fine}: The training corpus consists of 1.5M sentences sampled from $\sim$780k Wikipedia articles. 434 news report sentences are manually annotated using 113 types (2-level hierarchy) to form the test data;
(2) \textbf{OntoNotes}~\cite{weischedel2011ontonotes}: It has 13,109 news documents where 77 test documents are manually annotated using 89 types (3-level hierarchy)~\cite{gillick2014context};
(3) \textbf{BBN}~\cite{weischedel2005bbn}: It consists of 2,311 Wall Street Journal articles ($\sim$48k sentences) which are manually annotated using 93 types (2-level hierarchy).
Statistics of the datasets are shown in Table~\ref{table:data_stats}.

\smallskip
\noindent
\textsf{\textbf{Automatically Labeled Training Corpora.}}
We followed the process introduced in~\cite{ling2012fine} to generate training data for Wiki dataset.
%
For BBN and OntoNotes datasets, we utilized DBpedia Spotlight\footnote{\small\url{http://spotlight.dbpedia.org/}}, a state-of-the-art entity linking tool, to identify entity mentions from text and map them to Freebase entries. We then applied the types induced from Freebase to each entity mention and map them to the target types.
In particular, we discarded types which cannot be mapped to Freebase types in BBN dataset (47 out of 93).

\begin{table}[t]
\vspace{-0.0cm}
\begin{small}
\begin{center}
\vspace{0.0cm}
\begin{tabularx}{\linewidth}{x l l l}
\hline
\textbf{Data sets} & \textbf{Wiki} & \textbf{OntoNotes} & \textbf{BBN} \\
\hline
$\#$Types & 113 & 89 & 47 \\ 
$\#$Documents & 780,549 & 13,109 & 2,311 \\ 
$\#$Sentences & 1.51M & 143,709 & 48,899 \\ 
$\#$Training mentions & 2.69M & 223,342 & 109,090 \\ 
$\#$Ground-truth mentions & 563 & 9,604 & 121,001 \\ 
$\#$Features & 644,860 & 215,642 & 125,637\\
$\#$Edges in graph & 87M & 5.9M & 2.9M \\
\hline
\end{tabularx}
\vspace{-0.2cm}
\caption{Statistics of the datasets.}
\label{table:data_stats}
\vspace{-0.3cm}
\end{center}
\end{small}
\end{table}

\smallskip
\noindent
\textsf{\textbf{Feature Generation.}}
Table~\ref{table:features} lists the set of features used in our experiments,  which are similar to those used in \cite{Yogatama2015embedding,ling2012fine} except for topics and ReVerb patterns.
We used a 6-word window to extract context unigrams and bigrams for each mention (3 words on the left and the right).
We applied the Stanford CoreNLP tool~\cite{manning2014stanford} to get POS tags and dependency structures.
The word clusters were derived for each corpus using the Brown clustering algorithm\footnote{\small\url{https://github.com/percyliang/brown-cluster}}.
We discarded features which occur only once in the corpus. 
The same kinds of features were used in both label noise reduction (Sec.~\ref{subsec:exp_label_noise_reduction}) and fine-grained entity typing (Sec.~\ref{subsec:exp_improve_typing}) experiments.

\smallskip
\noindent
\textsf{\textbf{Type Correlation Graphs.}}
We used 2015-06-30 Freebase dump\footnote{\small\url{https://developers.google.com/freebase/data}} (1.9B triples, 115M entities, 16,701 types) and collected $266$M entity-type facts (triples with ``\textit{type.instance}" as predicate). Given two target types, we mapped them to Freebase types and followed the procedure introduced in Sec.~\ref{subsec:graph} to compute their KB-based correlation score.

\smallskip
\noindent
\textsf{\textbf{Evaluation Sets.}}
For Wiki and OntoNotes datasets,
we used the provided training/test set partitions of the corpora.
Since BBN corpus is fully annotated, we followed a 80/20 ratio to partition it into training/test sets. In particular, test sets for label noise reduction (Sec.~\ref{subsec:exp_label_noise_reduction}) consist of mentions in the original test set which can also be linked to KB entities (241, 1,190 and 32,353 mentions for Wiki, OntoNotes and BBN datasets, resp.).
We further created a validation set by randomly sampling 10\% mentions from the test set and used the remaining mentions to form the evaluation set.

\label{subsec:experiment_setting}
\smallskip
\noindent
\textsf{\textbf{Compared Methods.}}
We compared the proposed method (PLE) with its variants which model parts of the hypotheses, and three pruning heuristics~\cite{gillick2014context}. Several state-of-the-art embedding methods and partial-label learning methods were also implemented (or tested using their published codes):
(1) \textbf{Sib}~\cite{gillick2014context}: removes siblings types associated with a mention. A mention is discarded if all its types are pruned; (2) \textbf{Min}~\cite{gillick2014context}: removes types that appear only once in the document;
(3) \textbf{All}~\cite{gillick2014context}: first performs Sib pruning then Min pruning;
(4) \textbf{DeepWalk}~\cite{perozzi2014deepwalk}: DeepWalk is an approach for embedding a homogeneous graph with binary edges.
We applied it to the heterogeneous graph $G$ by treating all nodes as if they had the same type;
(4) \textbf{LINE}~\cite{tang2015line}:
We used second-order LINE model and edge sampling algorithm on feature-type bipartite graph (edge weight $w_{jk}$ is the number of mentions having feature $f_j$ and type $y_k$);
(5) \textbf{WSABIE}~\cite{Yogatama2015embedding}: adopts WARP loss with kernel extension to learn embeddings of features and types;
(6) \textbf{PTE}~\cite{tang2015pte}:
We applied PTE joint training algorithm on subgraphs \begin{small}$G_{MF}$\end{small} and \begin{small}$G_{MY}$\end{small}.
(7) \textbf{PL-SVM}~\cite{nguyen2008classification}: Partial-label SVM uses a margin-based loss to handle label noise.
(8) \textbf{CLPL}~\cite{cour2011learning}: uses a linear model to encourage large average scores for candidate types. We adopted the suggested setting (SVM with square hinge loss).

For PLE, besides the proposed model, \textbf{PLE}, which adopts KB-based type correlation subgraph, we compare (1) \textbf{PLE-NoCo}: This variant does not consider type correlation subgraph \begin{small}$G_{YY}$\end{small} in the objective in Eq.~\eqref{eq:objective}; and (2) \textbf{PLE-CoH}: It adopts type hierarchy-based correlation subgraph.

\begin{table}
\begin{scriptsize}
\vspace{-0.0cm}
\begin{center}
\hspace*{-0.4cm}
\begin{tabularx}{89mm}{l | l XXl XXX}
\hline
 & & \multicolumn{3}{c}{\textbf{\textsf{Macro}}} &  \multicolumn{3}{c}{\textbf{\textsf{Micro}}}\\
\textbf{Method}
& \textbf{\textsf{Acc}}
& \textbf{\textsf{P}} & \textbf{\textsf{R}}  & \textbf{\textsf{F1}}
& \textbf{\textsf{P}} & \textbf{\textsf{R}} & \textbf{\textsf{F1}}
\\ \hline
Raw
& 0.513
& 0.735 & \textbf{0.844} & 0.785
& 0.687 & \textbf{0.850} & 0.760 \\
Sib
& 0.516
& 0.707 & 0.703 & 0.705
& 0.689 & 0.690 & 0.690 \\
Min
& 0.509
& 0.735 & 0.833 & 0.781
& 0.688 & 0.838 & 0.756 \\
All
& 0.509
& 0.709 & 0.699 & 0.704
& 0.690 & 0.686 & 0.688 \\
DeepWalk-Raw
& 0.545
& 0.676 & 0.631 & 0.652
& 0.663 & 0.647 & 0.655 \\
LINE-Raw
& 0.703
& 0.766 & 0.753 & 0.759
& 0.771 & 0.768 & 0.770 \\
WSABIE-Raw
& 0.713
& 0.776 & 0.766 & 0.771
& 0.802 & 0.783 & 0.766 \\
PTE-Raw
& 0.703
& 0.824 & 0.775 & 0.799
& 0.833 & 0.773 & 0.802 \\
\hline
PLE-NoCo
& 0.755
& 0.829 & 0.814 & 0.821
& 0.836 & 0.822 & 0.829 \\
PLE-CoH
& 0.788
& 0.851 & 0.837 & 0.844
& 0.846 & 0.840 & 0.843 \\
PLE
& \textbf{0.812}
& \textbf{0.888} & 0.840 & \textbf{0.863}
& \textbf{0.883} & \textbf{0.850} & \textbf{0.867} \\
\hline
\end{tabularx}
\vspace{-0.2cm}
\caption{Performance comparisons on LNR on \textbf{BBN} dataset.}
\label{table:reduce_label_noise_bbn}
\end{center}
\vspace{-0.6cm}
\end{scriptsize}
\end{table}

\begin{table*}
\begin{scriptsize}
\vspace{-0.2cm}
\begin{center}
\begin{tabularx}{\textwidth}{  l | l XXl XXl | l XXl XXl }
\hline
 & \multicolumn{7}{c|}{\textbf{Wiki}} &  \multicolumn{7}{c}{\textbf{OntoNotes}}\\
\textbf{Method}
& \textbf{\textsf{Acc}}
& \textbf{\textsf{Ma-P}} & \textbf{\textsf{Ma-R}}  & \textbf{\textsf{Ma-F1}}
& \textbf{\textsf{Mi-P}} & \textbf{\textsf{Mi-R}} & \textbf{\textsf{Mi-F1}}
& \textbf{\textsf{Acc}}
& \textbf{\textsf{Ma-P}} & \textbf{\textsf{Ma-R}}  & \textbf{\textsf{Ma-F1}}
& \textbf{\textsf{Mi-P}} & \textbf{\textsf{Mi-R}} & \textbf{\textsf{Mi-F1}}
\\ \hline
Raw
& 0.373
& 0.558 & \textbf{0.681} & 0.614
& 0.521 & \textbf{0.719} & 0.605
& 0.480
& 0.671 & \textbf{0.793} & 0.727
& 0.576 & \textbf{0.786} & 0.665 \\
Sib~\cite{gillick2014context}
& 0.373
& 0.583 & 0.636 & 0.608
& 0.578 & 0.653 & 0.613
& 0.487
& 0.710 & 0.732 & 0.721
& 0.675 & 0.702 & 0.688 \\
Min~\cite{gillick2014context}
& 0.373
& 0.561 & 0.679 & 0.615
& 0.524 & 0.717 & 0.606
& 0.481
& 0.680 & 0.777 & 0.725
& 0.592 & 0.763 & 0.667 \\
All~\cite{gillick2014context}
& 0.373
& 0.585 & 0.634 & 0.608
& 0.581 & 0.651 & 0.614
& 0.487
& 0.716 & 0.724 & 0.720
& 0.686 & 0.691 & 0.689 \\
DeepWalk-Raw~\cite{perozzi2014deepwalk}
& 0.328
& 0.598 & 0.459 & 0.519
& 0.595 & 0.367 & 0.454
& 0.441
& 0.625 & 0.708 & 0.664
& 0.598 & 0.683 & 0.638 \\
LINE-Raw~\cite{tang2015line}
& 0.349
& 0.600 & 0.596 & 0.598
& 0.590 & 0.610 & 0.600
& 0.549
& 0.699 & 0.770 & 0.733
& 0.677 & 0.754 & 0.714 \\
WSABIE-Raw~\cite{Yogatama2015embedding}
& 0.332
& 0.554 & 0.609 & 0.580
& 0.557 & 0.633 & 0.592
& 0.482
& 0.686 & 0.743 & 0.713
& 0.667 & 0.721 & 0.693 \\
PTE-Raw~\cite{tang2015pte}
& 0.419
& 0.678 & 0.597 & 0.635
& 0.686 & 0.607 & 0.644
& 0.529
& 0.687 & 0.754 & 0.719
& 0.657 & 0.733 & 0.693 \\
\hline
PLE-NoCo
& 0.556
& 0.795 & 0.678 & 0.732
& 0.804 & 0.668 & 0.730
& 0.593
& 0.768 & 0.773 & 0.770
& 0.751 & 0.762 & 0.756 \\
PLE-CoH
& 0.568
& 0.805 & 0.671 & 0.732
& 0.808 & 0.704 & 0.752
& 0.620
& 0.789 & 0.785 & 0.787
& 0.778 & 0.769 & 0.773 \\
PLE
& \textbf{0.589}
& \textbf{0.840} & 0.675 & \textbf{0.749}
& \textbf{0.833} & 0.705 & \textbf{0.763}
& \textbf{0.639}
& \textbf{0.814} & 0.782 & \textbf{0.798}
& \textbf{0.791} & 0.766 & \textbf{0.778} \\
\hline
\end{tabularx}
\vspace{-0.2cm}
\caption{Performance comparisons on LNR on \textbf{Wiki} and \textbf{OntoNotes} datasets.}
\label{table:reduce_label_noise_wiki_ontonotes}
\end{center}
\vspace{-0.3cm}
\end{scriptsize}
\end{table*}

\smallskip
\noindent
\textsf{\textbf{Parameter Settings.}}
In our testing of PLE and its variants, we set \begin{small}$\alpha=0.25$\end{small}, \begin{small}$\eta=0.1$\end{small} and \begin{small}$\lambda=10^{-4}$\end{small} (see Fig.~\ref{figure:tune_lambda}) by default, based on the analysis on validation sets. For convergence criterion, we stopped the loop in Algorithm~\ref{algorithm:PLE} if the relative change of \begin{small}$\O$\end{small} in Eq.~\eqref{eq:objective} is smaller than \begin{small}$10^{-4}$\end{small}.
For fair comparison, the dimensionality of embeddings $d$ was set to $50$ and the number of negative samples (\begin{small}$Z$\end{small} in PLE) was set to $5$ for PLE, PTE and LINE, as used in \cite{tang2015line}.
For DeepWalk, we set window size as 10, walk length as 40, walks per vertex as 40, as used in~\cite{perozzi2014deepwalk}.
Learning rates of LINE and PTE were set to \begin{small}$\rho_t=\rho_0(1-t/T)$\end{small} with \begin{small}$\rho_0=0.025$\end{small} where $T$ is total number of edge samples (set to 10 times of the number of edges), as used in \cite{tang2015pte} and \cite{tang2015line}.
After tuning on validation sets, we set learning rate as $ 0.001$ for WSABIE, and set the regularization parameters in PL-SVM and CLPL as $0.1$.

\smallskip
\noindent
\textsf{\textbf{Evaluation Metrics.}}
We use F1 score computed from Precision and Recall scores in 3 different granualities~\cite{ling2012fine,Yogatama2015embedding}.
Let \begin{small}$\P$\end{small} denote evaluation set. For mention \begin{small}$m\in\P$\end{small}, we denote its ground-truth types as $t_m$ and the predicted types as $\widehat{t}_m$.
\vspace{-0.4cm}
\begin{itemize}[leftmargin=10pt]\itemsep-0.1cm
\item \textbf{Strict}: The prediction is considered correct if and only if \begin{small}$t_m = \widehat{t}_m$\end{small}: 
Accuracy (\textbf{\textsf{Acc}}) = \begin{small}$\sum_{m\in\P}\one(t_m = \widehat{t}_m)/|\P|$\end{small}.
\item \textbf{Loose Macro}: The Macro-Precision (\textbf{\textsf{Ma-P}}) and Macro-Recall (\textbf{\textsf{Ma-R}}) are computed for each mention: Ma-P = \begin{small}$\frac{1}{|\P|}\sum\limits_{m\in\P}|t_m \cap \widehat{t}_m|/|\widehat{t}_m|$\end{small} and Ma-R = \begin{small}$\frac{1}{|\P|}\sum\limits_{m\in\P}|t_m \cap \widehat{t}_m|/|t_m|$\end{small}.
\item \textbf{Loose Micro}: The Micro-Precision (\textbf{\textsf{Mi-P}}) and Micro-Recall (\textbf{\textsf{Mi-R}}) scores are averages over all mentions, \ie, Mi-P = \begin{small}$\frac{\sum_{m\in\P}|t_m \cap \widehat{t}_m|}{\sum_{m\in\P}|\widehat{t}_m|}$\end{small} and Mi-R = \begin{small}$\frac{\sum_{m\in\P}|t_m \cap \widehat{t}_m|}{\sum_{m\in\P}|t_m|}$\end{small}.
\end{itemize}

\subsection{Label Noise Reduction}
\label{subsec:exp_label_noise_reduction}
We first conduct \textit{intrinsic evaluation} on how accurately PLE and the other methods can estimate the true types of mentions (\ie, \begin{small}$\{\Y_i^*\}$\end{small}) from its noisy candidate type set (\ie, \begin{small}$\{\Y_i\}$\end{small}).
Let \begin{small}$\P_L$\end{small} denote the test mentions which can be linked to KB.
We evaluate the quality of the candidate type set (\ie, Raw), and three pruning methods on \begin{small}$\P_L$\end{small}.
For PLE and other embedding methods,
we learn models on \begin{small}$\D\cup\P_L$\end{small} using the candidate types, and evaluate the estimation results on the ground-truth types of \begin{small}$\P_L$\end{small}.
To test pruning methods, we further apply them on \begin{small}$\D\cup\P_L$\end{small} (the pruned corpus is denoted as \begin{small}$\D_P$\end{small}), and learn the compared embedding models on \begin{small}$\D_P$\end{small}.

\smallskip
\noindent
\textsf{\textbf{1. Comparing PLE with the other methods.}} Tables~\ref{table:reduce_label_noise_bbn} and \ref{table:reduce_label_noise_wiki_ontonotes} summarize the comparison results on the three datasets.
For embedding models learned on different pruned corpora, we only show the combination that yields the best result.
Overall, PLE and its variants outperform others on Accuracy, Precision and F1 scores, and achieve Recall close to that of Raw---Raw's Recall is the upper bound since type inference is conducted within the candidate type set. In particular, PLE obtains a 40.57\% improvement in Accuracy and 23.89\% improvement in Macro-Precision compared to the best baseline PTE-Raw on Wiki dataset, and improves Accuracy by 16.39\% compared to the best baseline LINE-Raw, on the OntoNotes dataset.
All three pruning methods suffer from low Recall because they  filter conflicting subtypes (\eg, Sib) and/or infrequent types (\eg, Min) aggressively. 
Superior performance of PLE demonstrates the needs of LNR to identify true types from the candidate type sets (versus aggressive type deletion).
PTE utilizes heterogeneous graph structure but suffers from low Precision and Recall, since it does not handle the noisy mention-candidate type links and does not model type correlation. PLE's performance improvement validates Hypotheses \ref{hypothesis:partial_label} and \ref{hypothesis:type_correlation}.
Both WSABIE and LINE aggregate feature-mention-type associations into feature-type associations to reduce the effect of noisy types, but statistics of infrequent features may be biased due to noisy mention-type links. PLE obtains superior performance because it effectively models the noisy type labels.

\smallskip
\noindent
\textsf{\textbf{2. Comparing PLE with its variants.}}
Comparing with PLE-NoCo, PLE gains performance from capturing type semantic similarity with the type correlation subgraph \begin{small}$G_{YY}$\end{small}, which assists in embedding rare types in the corpus. PLE always outperforms PLE-CoH on all metrics on the three datasets. The enhancement mainly comes from modeling type correlation with entity-type facts in KB, which yields more accurate and complete type correlation statistics compared to the type hierarchy-based approach (see Sec.~\ref{subsec:graph}).

\begin{table}[h]
\begin{scriptsize}
\vspace{-0.0cm}
\begin{tabularx}{\linewidth}{| m{1.2cm}|X|X|}
\hline
\textbf{Text}
&  \textit{\textbf{NASA}} says it may decide by tomorrow whether another space walk will be needed ...
& ... the board of \textit{directors} which are composed of twelve members directly appointed by the \textit{\textbf{Queen}}.
\\ \hline
\textbf{Wiki Page}
& \url{https://en.wikipedia.org/wiki/NASA}
& \url{https://en.wikipedia.org/wiki/Elizabeth_II}
\\ \hline
\textbf{Cand. type set}
& \texttt{person}, \texttt{artist}, \texttt{location}, \texttt{structure}, \texttt{organization}, \texttt{company}, \texttt{news\_company}
& \texttt{person}, \texttt{artist}, \texttt{actor}, \texttt{author}, \texttt{person\_title}, \texttt{politician}
\\ \hline
\textbf{WSABIE}
& {\color{red!70}\texttt{person}}, {\color{red!70}\texttt{artist}}
& \texttt{person}, {\color{red!70}\texttt{artist}}
\\ \hline
\textbf{PTE}
& \texttt{organization}, \texttt{company}, {\color{red!70}\texttt{news\_company}}
& \texttt{person}, {\color{red!70}\texttt{artist}}
\\ \hline
\textbf{PLE}
& \texttt{organization}, \texttt{company}
& \texttt{person}, \texttt{person\_title}
\\ \hline
\end{tabularx}
\vspace{-0.3cm}
\caption{Example output of PLE and the compared methods on two news sentences from the \textbf{OntoNotes} dataset.}
\label{table:example_output}
\vspace{-0.2cm}
\end{scriptsize}
\end{table}

\smallskip
\noindent
\textsf{\textbf{3. Example output on news articles.}}
Table~\ref{table:example_output} shows the types estimated by PLE, PTE and WSABIE on three news sentences from OntoNotes dataset: PLE predicts fine-grained types with better accuracy (\eg, \texttt{person\_title}) and avoids from overly-specific predictions (\eg, \texttt{news\_company}).

\begin{figure}[h]
\centering
\vspace{-0.0cm}
\subfigure[Effect of training set size]{
\includegraphics[width = 39.5 mm]{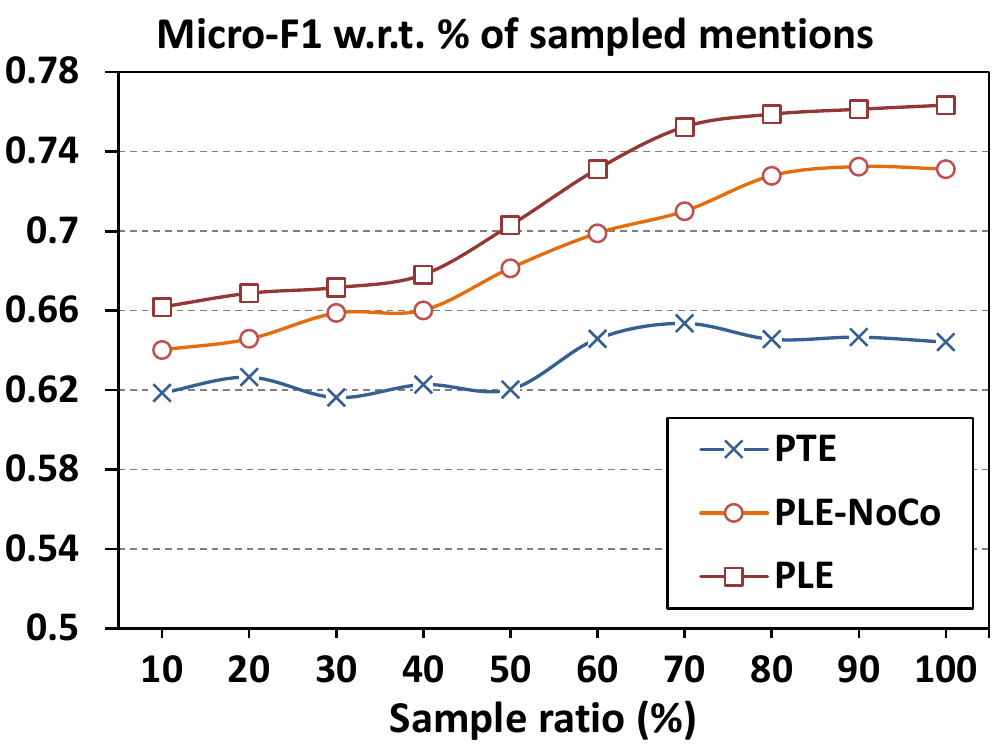}
\label{figure:train_sample}
}
\subfigure[Performance change w.r.t. $\lambda$]{
\includegraphics[width = 39.5 mm]{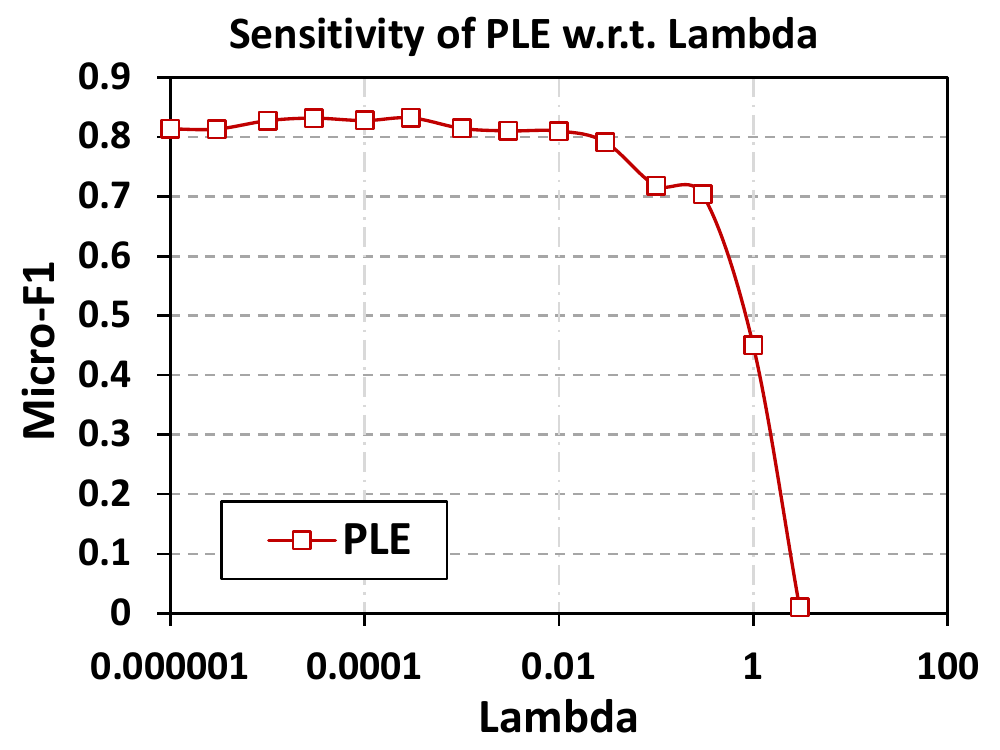}
\label{figure:tune_lambda}
}
\vspace{-0.4cm}
\caption{Performance change with respect to (a) sampling ratio of mentions from the training set on the Wiki dataset; and (b) regularization parameter $\lambda$ on the BBN dataset.}
\vspace{-0.2cm}
\end{figure}

\smallskip
\noindent
\textsf{\textbf{4. Testing the effect of training set size.}}
Experimenting with the same settings for graph construction and model learning, Fig.~\ref{figure:train_sample} shows the performance trend on Wiki dataset when varying the sampling ratio (subset of mentions randomly sampled from the training set \begin{small}$\D$\end{small}). Performance of all methods improves as the ratio increases, and becomes insensitive as the ratio~\begin{small}$>70$\%\end{small}. PLE always outperforms its variant and the best baseline PTE. In particular, PLE model trained at $10$\% sampling rate outperforms the best PTE model (obtained at $70$\% sampling rate).

\smallskip
\noindent
\textsf{\textbf{5. Testing sensitivity of the tuning parameter.}} Fig.~\ref{figure:tune_lambda} analyzes the performance sensitivity of PLE with respect to $\lambda$---the only tuning parameter in the proposed model---on BBN dataset. Performance of PLE becomes insensitive as $\lambda$ becomes small enough (\ie, 0.01). We set $\lambda=10^{-4}$ throughout our experiments for PLE and its variants.

\begin{table*}
\begin{scriptsize}
\vspace{-0.3cm}
\vspace{0.0cm}
\begin{center}
\begin{tabularx}{\textwidth}{ l |  l | XXX | XXX | XXX }
\hline
\textbf{Typing} & \textbf{Noise Reduction} & \multicolumn{3}{c|}{\textbf{Wiki}} &  \multicolumn{3}{c|}{\textbf{OntoNotes}} & \multicolumn{3}{c}{\textbf{BBN}}\\
\textbf{System} & \textbf{Method}
& \textbf{\textsf{Acc}} & \textbf{\textsf{Ma-F1}}  & \textbf{\textsf{Mi-F1}}
& \textbf{\textsf{Acc}} & \textbf{\textsf{Ma-F1}}  & \textbf{\textsf{Mi-F1}}
& \textbf{\textsf{Acc}} & \textbf{\textsf{Ma-F1}}  & \textbf{\textsf{Mi-F1}}
\\ \hline
N/A & PL-SVM~\cite{nguyen2008classification}
& 0.428 & 0.613 & 0.571
& 0.465 & 0.648 & 0.582
& 0.497 & 0.679 & 0.677 \\
N/A & CLPL~\cite{cour2011learning}
& 0.162 & 0.431 & 0.411
& 0.438 & 0.603 & 0.536
& 0.486 & 0.561 & 0.582 \\
\hline
& Raw
& 0.288 & 0.528 & 0.506
& 0.249 & 0.497 & 0.446
& 0.523 & 0.576 & 0.587 \\
& Min~\cite{gillick2014context}
& 0.325 & 0.566 & 0.536
& 0.295 & 0.523 & 0.470
& 0.524 & 0.582 & 0.595 \\
& All~\cite{gillick2014context}
& 0.417 & 0.591 & 0.545
& 0.305 & 0.552 & 0.495
& 0.495 & 0.563 & 0.568 \\
\textbf{HYENA}~\cite{yosef2012hyena}
& WSABIE-Min~\cite{Yogatama2015embedding}
& 0.199 & 0.462 & 0.459
& 0.400 & 0.565 & 0.521
& 0.524 & 0.610 & 0.621 \\
& PTE-Min~\cite{tang2015pte}
& 0.238 & 0.542 & 0.522
& 0.452 & 0.626 & 0.572
& 0.545 & 0.639 & 0.650 \\
& PLE-NoCo
& 0.517 & 0.672 & 0.634
& 0.496 & 0.658 & 0.603
& 0.650 & 0.709 & 0.703 \\
& PLE
& \textbf{0.543} & \textbf{0.695} & \textbf{0.681}
& \textbf{0.546} & \textbf{0.692} & \textbf{0.625}
& \textbf{0.692} & \textbf{0.731} & \textbf{0.732} \\
\hline
& Raw
& 0.474 & 0.692 & 0.655
& 0.369 & 0.578 & 0.516
& 0.467 & 0.672 & 0.612 \\
& Min
& 0.453 & 0.691 & 0.631
& 0.373 & 0.570 & 0.509
& 0.444 & 0.671 & 0.613 \\
& All
& 0.453 & 0.648 & 0.582
& 0.400 & 0.618 & 0.548
& 0.461 & 0.636 & 0.583 \\
\textbf{FIGER}~\cite{ling2012fine}
& WSABIE-Min
& 0.455 & 0.646 & 0.601
& 0.425 & 0.603 & 0.546
& 0.481 & 0.671 & 0.618 \\
& PTE-Min
& 0.476 & 0.670 & 0.635
& 0.494 & 0.675 & 0.618
& 0.513 & 0.674 & 0.657 \\
& PLE-NoCo
& 0.543 & 0.726 & 0.705
& 0.547 & 0.699 & 0.639
& 0.643 & 0.753 & 0.721 \\
& PLE
& \textbf{0.599} & \textbf{0.763} & \textbf{0.749}
& \textbf{0.572 }& \textbf{0.715} & \textbf{0.661}
& \textbf{0.685} & \textbf{0.777} & \textbf{0.750} \\
\hline
\end{tabularx}
\vspace{-0.2cm}
\caption{Study of performance improvement on fine-grained typing systems \textbf{FIGER}~\cite{ling2012fine} and \textbf{HYENA}~\cite{yosef2012hyena} on the three datasets.}
\label{table:improve_typing}
\end{center}
\vspace{-0.2cm}
\end{scriptsize}
\end{table*}

\subsection{Fine-Grained Entity Typing}
\label{subsec:exp_improve_typing}
We further conduct \textit{extrinsic evaluation} on fine-grained typing to study the performance gain from denoising the automatically generated training corpus \begin{small}$\D$\end{small}.
Two state-of-the-art fine-grained type classifiers, HYENA~\cite{yosef2012hyena} and FIGER~\cite{ling2012fine}, are trained using the same set of features as listed in Table~\ref{table:features} on the denoised corpus (denoted as \begin{small}$\D_d$\end{small}), which is generated using PLE or the other compared methods.
Trained classifiers are then tested on the evaluation set \begin{small}$\P$\end{small}.
Similar to the process introduced in Sec.~\ref{subsec:exp_label_noise_reduction}, embedding models trained on different pruned corpora are compared as well.
We also compare with partial-label learning methods PL-SVM~\cite{nguyen2008classification} and CLPL~\cite{cour2011learning}, which are trained on \begin{small}$\D$\end{small} and evaluated on \begin{small}$\P$\end{small}.

\smallskip
\noindent
\textsf{\textbf{1. Comparing with the other noise reduction methods.}}
Table~\ref{table:improve_typing} reports the comparison results of the two best performing pruning methods and embedding methods on the three datasets.
Both typing systems achieve superior performance on all metrics when using PLE and its variant to denoise the training corpus. In particular, PLE improves FIGER's Accuracy (\ie, Raw) by 33.53\% and HYENA's Accuracy by 26.97\% on the BBN dataset. Compared to the best baseline PTE-Min, PLE obtains over 28\% improvement in HYENA's F1 scores and over 13\% enhancement in FIGER's F1 scores on the Wiki dataset. Superior performance of PLE demonstrates the effectiveness of the proposed margin-based loss in modeling noisy candidate types.
PLE always outperforms PLE-NoCo on all metrics on both typing systems. It gains performance from capturing type correlation, by jointly modeling the type-type links in the embedding process.
In particular, we observe that pruning methods do not always improve the performance (\eg, ``All" pruning results in a 11.15\% drop in Macro-F1 score on FIGER on the Wiki dataset), since they aggressively filter out subtypes and/or rare types in the corpus, which may lead to low Recall.

\smallskip
\noindent
\textsf{\textbf{2. Comparing with partial-label learning methods.}} Comparing with PL-SVM and CLPL, both typing systems obtain superior performance when PLE is applied to denoise the training corpora. PL-SVM adopts a modified margin-based objective to fit linear models on features using the noisy candidate types, but it assume that only \textit{one} candidate type is correct and does not consider semantic similarity between the types. CLPL simply averages the model output for all candidate types, and thus may generate results biased to frequent false types. Superior performance of PLE mainly comes from jointly modeling of type correlation derived from KB and feature-mention co-occurrences in the corpus.

\begin{figure}[h]
\vspace{-0.0cm}
\includegraphics[width = 81 mm]{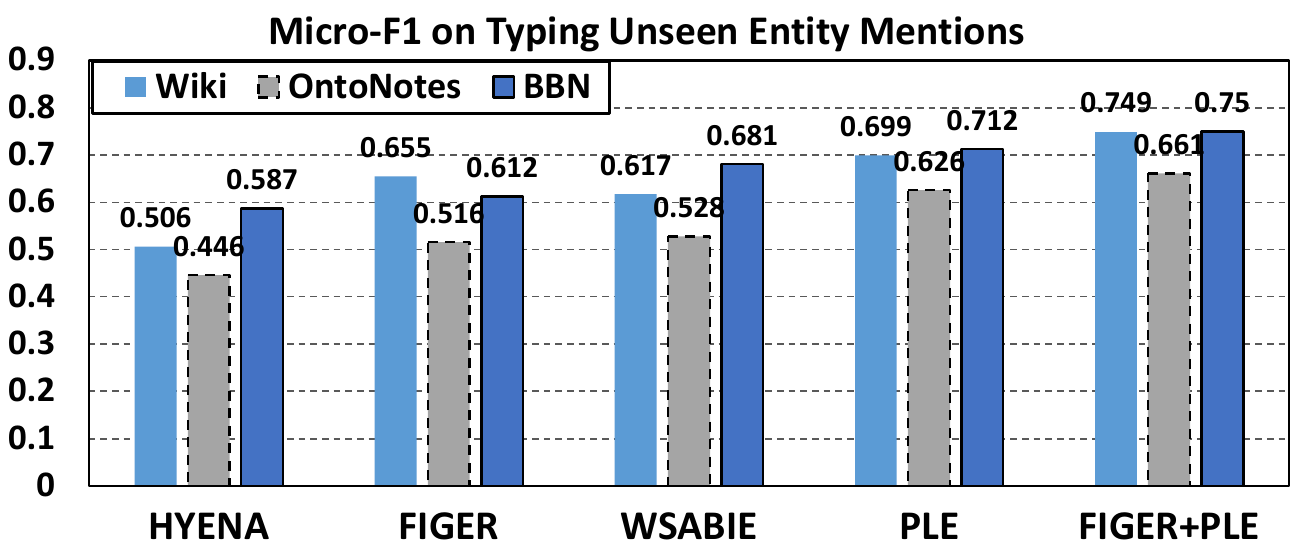}
\vspace{-0.3cm}
\caption{Performance comparison in terms of Micro-F1 for predicting types of unseen entity mentions in the three datasets.}
\label{figure:typing_unseen}
\vspace{-0.2cm}
\end{figure}

\smallskip
\noindent
\textsf{\textbf{3. Testing on unseen mentions.}}
Fig.~\ref{figure:typing_unseen} compares PLE with the other methods for predicting types of unseen mentions in the three datasets. We used the learned feature embeddings and type embeddings to estimate the type-path for each mention in \begin{small}$\P$\end{small}.
PLE outperforms both FIGER and HYENA systems (\eg, over 21\% improvement in Micro-F1 on the OntoNotes dataset)---demonstrating the predictive power of the learned embeddings, and the effectiveness of modeling noisy candidate types. Although FIGER trained on PLE-denoised corpus obtains superior F1 scores, PLE can achieve competitive performance without training an additional classifier (\ie, more efficiently).

\subsection{Case Analyses}
\begin{figure}[h]
\centering
\vspace{-0.0cm}
\subfigure[Test at different type levels]{
\includegraphics[width = 39.5 mm]{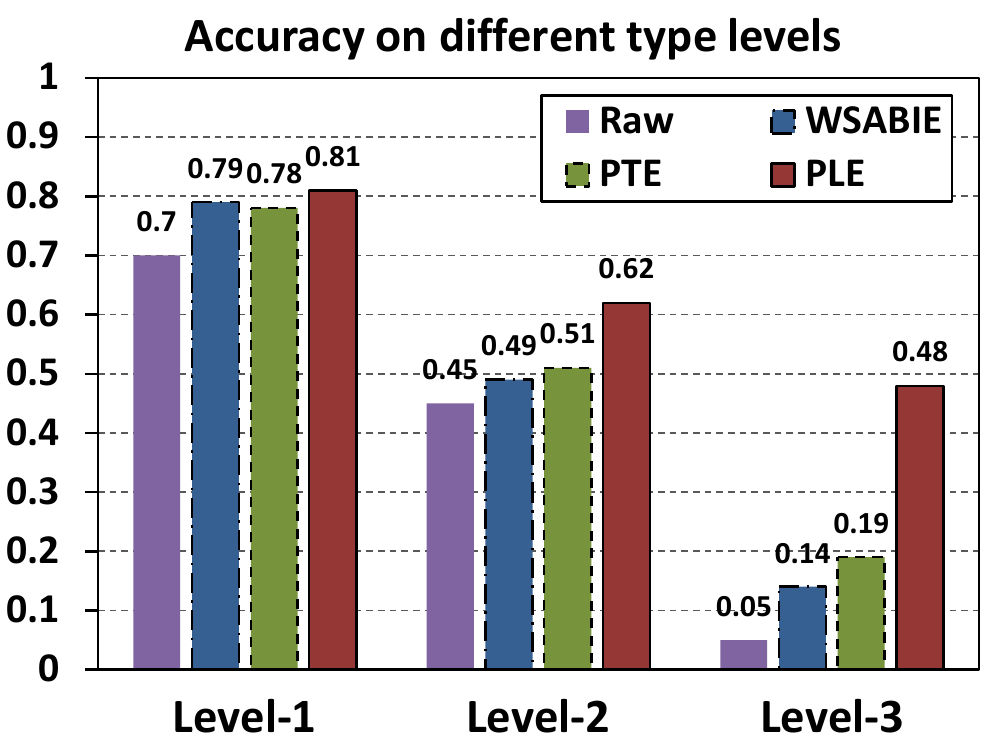}
\label{figure:type_level}
}
\subfigure[Iterative Re-training]{
\includegraphics[width = 39.5 mm]{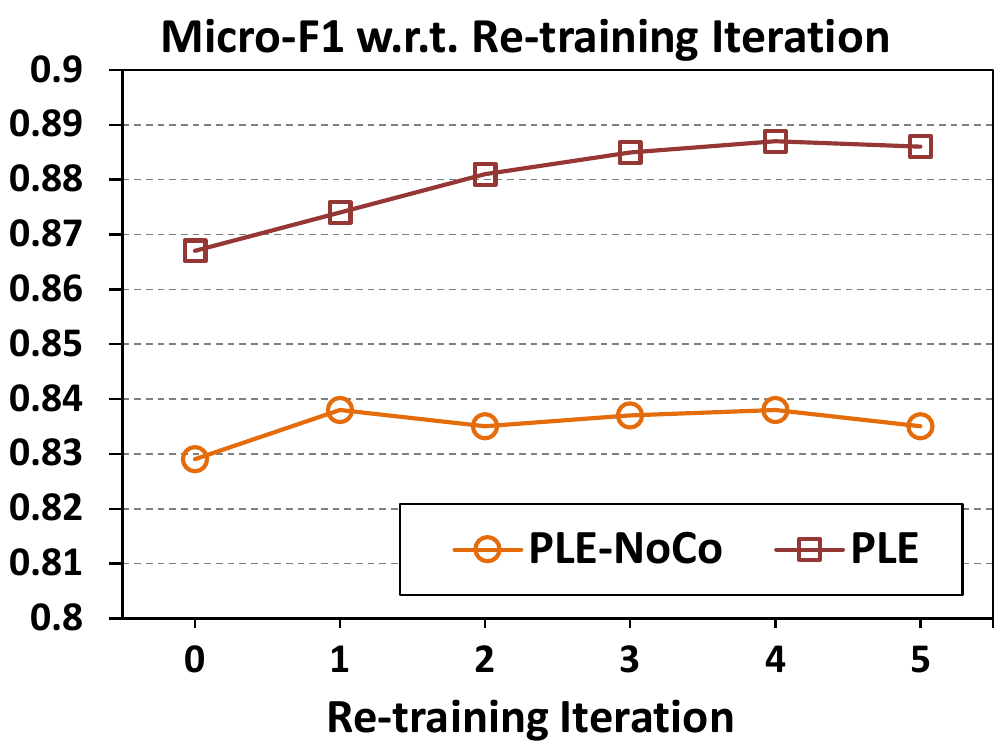}
\label{figure:boostrap}
}
\vspace{-0.4cm}
\caption{Performance change of PLE (a) at different levels of the type hierarchy on the OntoNotes dataset; and (b) with respect to the number of re-training iterations on the BBN dataset.}
\vspace{-0.2cm}
\end{figure}

\noindent
\textsf{\textbf{1. Testing at different type levels.}}
Fig.~\ref{figure:type_level} reports the Accuracy of PLE, PTE and WSABIE on recovering ground-truth types at different levels of the target type hierarchy \begin{small}$\Y$\end{small}. The results shows that it is more difficult to distinguish among deeper (more fine-grained) types. PLE always outperforms the other two method, and achieves a \textbf{153\%} improvement in Accuracy, compared to the best baseline PTE on level-3 types. The gain mainly comes from explicitly modeling the noisy candidate types, since most mention-type links on fine-grained types are false positives.

\smallskip
\noindent
\textsf{\textbf{2. Iterative re-training of PLE.}} We re-train PLE model and its variants using the corpus \begin{small}$\D_d$\end{small} which has been denoised by PLE, to analyze the effect of boostrapping PLE. To avoid overly-low Recall, in each iteration we conduct type inference in the original candidate type set \begin{small}$\{\Y_i\}$\end{small}.
Fig.~\ref{figure:boostrap} shows that the performance gain becomes marginal after 3 iterations of re-training. This may be because the learned embeddings in the first round of training already capture all the signals encoded in the heterogeneous graph---the updated mention-type subgraph from the denoised corpus does not cause significant changes to the embeddings.

\section{Related Work}
\label{sec:related}
\noindent \textsf{\textbf{Fine-Grained Entity Typing.}}
There have been extensive studies on entity recognition and typing. In terms of the dependence on context information, existing work can be categorized into context-dependent~\cite{nadeau2007survey,ling2012fine} and context-independent approaches~\cite{nakashole2013fine,lin2012no}. Work along both lines can be further categorized in terms of the type granularity that is considered.
Traditional named entity recognition systems~\cite{manning2014stanford} focus on coarse types (\eg, \texttt{person}, \texttt{location}) and cast the problem as multi-class classification following the type mutual exclusion assumption (\ie, one type per mention)~\cite{nadeau2007survey}.
Recent work has focused on a much larger set of fine-grained types~\cite{yosef2012hyena,ling2012fine}. As type mutual exclusion assumption no longer holds, they cast the problem as multi-label multi-class (hierarchical) classification problems~\cite{gillick2014context,yosef2012hyena,ling2012fine}, or make use of various supervised embedding techniques~\cite{Yogatama2015embedding,Dong2015HybridNeural} to jointly derive feature representations in classification tasks.

Most existing fine-grained typing systems use distant supervision to generate training examples and assume that all candidate types so generated are correct. 
By contrast, our framework instead seeks to remove false positives, denoising the data and leaving only the correct ones for each mention based on its local context. Output of our task, \ie, denoised training data, helps train more effective classifiers for entity typing.
Gillick~\etal~\cite{gillick2014context} discuss the label noise issue in fine-grained typing and propose three type pruning heuristics. However, these pruning methods aggressively filter training examples and may suffer from low recall (see Table.~\ref{table:improve_typing}).

In the context of distant supervision, label noise issue has been studied for other information extraction tasks such as relation extraction~\cite{takamatsu2012reducing} and slot filling~\cite{ji2014tackling}. However, the form of supervision is different from that in entity typing.

\smallskip
\noindent \textsf{\textbf{Partial Label Learning.}}
Partial label learning (PLL)~\cite{zhang2014disambiguation,nguyen2008classification,cour2011learning} deals with the problem where each training example is associated with a set of candidate labels, where \textit{only one is correct}.
One intuitive strategy to solve the problem is to assume equal contribution of each candidate label and average the outputs from all candidate labels for prediction~\cite{cour2011learning}.
Another strategy is to treat true label as latent variable and  optimize objectives such as maximum likelihood criterion~\cite{liu2012conditional} and maximum margin criterion~\cite{nguyen2008classification} by EM procedure.


Unlike this PLL formulation, our problem can be seen as \textit{hierarchical classification with partial labels}. Existing PLL methods model a single true label for each training example and do not consider label correlation information.
We compare with simple extensions of PL-SVM~\cite{nguyen2008classification} and CLPL~\cite{cour2011learning} by applying the learned partial-label classifiers to predicted type-paths in a top-down manner (see Table.~\ref{table:improve_typing}).


\smallskip
\noindent \textsf{\textbf{Text and Network Embedding.}}
The proposed PLE framework incorporate embedding techniques used in modeling text data~\cite{Yogatama2015embedding,Dong2015HybridNeural,mikolov2013distributed}, and networks/graphs~\cite{tang2015pte,perozzi2014deepwalk,he2004locality}.
However, existing methods assume links are all correct (unsupervised) or labels are all true (supervised)---our approach seeks to \textit{delete noisy links and lables} in the embedding process. We compare with several embedding methods like PTE~\cite{tang2015line} to validate the proposed Hypothesis~\ref{hypothesis:partial_label} on noisy labels (see Sec.~\ref{subsec:exp_label_noise_reduction}).
With respect to modeling type correlation, our work is related to KB embedding~\cite{hu2015hiearchy,bordes2013translating}, which focuses on embedding \textit{global} information of the KB elements (\eg, entities, relations, types) into a low-dimensional space, although ours incorporates \textit{local} context information of entity mentions in text, and models KB-based type correlation jointly.

\vspace*{-0.2cm}
\section{Conclusion and Future Work}
\label{sec:conclusion}

We study a new task on reducing label noise in distant supervision for fine-grained entity typing, and propose a heterogeneous partial-label embedding framework (PLE) to denoise candidate types in automatically labeled training corpora.
Experiment results demonstrate that the proposed method can recover true type labels effectively and robustly, and the denoised training data can significantly enhance performance of state-of-the-art typing systems.
Interesting future work includes extending PLE's similarity function to model hierarchical type dependency~\cite{hu2015hiearchy}, deploying multi-sense embedding to model topics of contexts~\cite{Yogatama2015embedding}, and exploiting relation facts in KB jointly~\cite{bordes2013translating}. Embeddings learned by PLE can be directly used to predict types for unseen mentions, which saves time otherwise needed to build additional classifiers.
PLE is general and can be used to denoise training data in other domains (\eg, image annotation~\cite{weston2011wsabie}). 
\vspace*{-0.2cm}

\small
\bibliographystyle{abbrv}
\bibliography{16-FineType}

\end{document}